\documentclass[journal]{IEEEtran}

\usepackage{graphicx} 
\usepackage{bm}
\usepackage{epstopdf}
\usepackage{booktabs} 
\usepackage{multirow}
\usepackage{subfigure}
\usepackage{array}
\usepackage{algpseudocode}
\usepackage{amsmath}
\usepackage{amsfonts}
\usepackage{caption}
\usepackage{algorithm}
\usepackage{algpseudocode}
\usepackage{amsmath}
\usepackage{stfloats}
\usepackage{balance}
\ifCLASSOPTIONcompsoc
  \usepackage[nocompress]{cite}
\else
  \usepackage{cite}
\fi
\ifCLASSINFOpdf
\else
\fi
\begin{document}

\title{Graph Learning: A Survey}

\author{Feng~Xia,~\IEEEmembership{Senior Member,~IEEE,}
        Ke~Sun, Shuo~Yu,~\IEEEmembership{Member,~IEEE,} Abdul~Aziz,
        Liangtian~Wan,~\IEEEmembership{Member,~IEEE,}
        Shirui Pan,
        and~Huan~Liu,~\IEEEmembership{Fellow,~IEEE}
\thanks{F. Xia is with School of Engineering, IT and Physical Sciences, Federation University Australia, Ballarat, VIC 3353, Australia}
\thanks{K. Sun, S. Yu, A. Aziz, and L. Wan are with School of Software, Dalian University of Technology, Dalian 116620, China.}
\thanks{S. Pan is with Faculty of Information Technology, Monash University, Melbourne, VIC 3800, Australia.}
\thanks{H. Liu is with School of Computing, Informatics, and Decision Systems Engineering, Arizona State University, Tempe, AZ 85281, USA.}
\thanks{Corresponding author: Feng Xia; e-mail: f.xia@ieee.org}}

\markboth{IEEE Transactions on Artificial Intelligence,~Vol.~00, No.~0, ~2021}%
{Xia \MakeLowercase{\textit{et al.}}: Graph Learning: A Survey}

\maketitle

\begin{abstract}	
Graphs are widely used as a popular representation of the network structure of connected data. Graph data can be found in a broad spectrum of application domains such as social systems, ecosystems, biological networks, knowledge graphs, and information systems. With the continuous penetration of artificial intelligence technologies, graph learning (i.e., machine learning on graphs) is gaining attention from both researchers and practitioners. Graph learning proves effective for many tasks, such as classification, link prediction, and matching. Generally, graph learning methods extract relevant features of graphs by taking advantage of machine learning algorithms. In this survey, we present a comprehensive overview on the state-of-the-art of graph learning. Special attention is paid to four categories of existing graph learning methods, including graph signal processing, matrix factorization, random walk, and deep learning. Major models and algorithms under these categories are reviewed respectively. We examine graph learning applications in areas such as text, images, science, knowledge graphs, and combinatorial optimization. In addition, we discuss several promising research directions in this field.

\end{abstract}

\begin{IEEEkeywords}
Graph learning, graph data, machine learning, deep learning, graph neural networks, network representation learning, network embedding.
\end{IEEEkeywords}

%
\IEEEpeerreviewmaketitle

\section*{Impact Statement}
Real-world intelligent systems generally rely on machine learning algorithms handling data of various types. Despite their ubiquity, graph data have imposed unprecedented challenges to machine learning due to their inherent complexity. Unlike text, audio and images, graph data are embedded in an irregular domain, making some essential operations of existing machine learning algorithms inapplicable. Many graph learning models and algorithms have been developed to tackle these challenges. This paper presents a systematic review of the state-of-the-art graph learning approaches as well as their potential applications. The paper serves multiple purposes. First, it acts as a quick reference to graph learning for researchers and practitioners in different areas such as social computing, information retrieval, computer vision, bioinformatics, economics, and e-commence. Second, it presents insights into open areas of research in the field. Third, it aims to stimulate new research ideas and more interests in graph learning.

\section{Introduction}\label{sec:introduction}

\IEEEPARstart{G}{raphs}, also referred to as networks, can be extracted from various real-world relations among abundant entities. Some common graphs have been widely used to formulate different relationships, such as social networks, biological networks, patent networks, traffic networks, citation networks, and communication networks~\cite{fortunato2018science,liu2020wos,liu2019shifu2}. A graph is often defined by two sets, i.e., vertex set and edge set. Vertices represent entities in  graph, whereas edges represent relationships between those entities. Graph learning has attracted considerable attention because of its wide applications in the real world, such as data mining and knowledge discovery. Graph learning methods have gained  increasing popularity for capturing complex relationships, as graphs exploit essential and relevant relations among vertices~\cite{zhang2018network,sun2020bigdata}. For example, in microblog networks, the spread trajectory of rumors can be tracked by detecting information cascades. In biological networks, new treatments for difficult diseases can be discovered by inferring protein interactions. In traffic networks, human mobility patterns can be predicted by analyzing the co-occurrence phenomenon with different timestamps~\cite{xia2020TITSstation}. Efficient analysis of these networks massively depends on the way how networks are represented.

\subsection{What is Graph Learning?}
Generally speaking, graph learning refers to machine learning on graphs. Graph learning methods map the features of a graph to feature vectors with the same dimensions in the embedding space. A graph learning model or algorithm directly converts the graph data into the output of the graph learning architecture without projecting the graph into a low dimensional space. Most graph learning methods are based on or generalized from deep learning techniques, because deep learning techniques can encode and represent graph data into vectors. The output vectors of graph learning are in continuous space. The target of graph learning is to extract the desired features of a graph. Thus the representation of a graph can be easily used by downstream tasks such as node classification and link prediction without an explicit embedding process. Consequently, graph learning is a more powerful and meaningful technique for graph analysis.

\begin{figure*}[htb]
	\centering
	\includegraphics[width=6in]{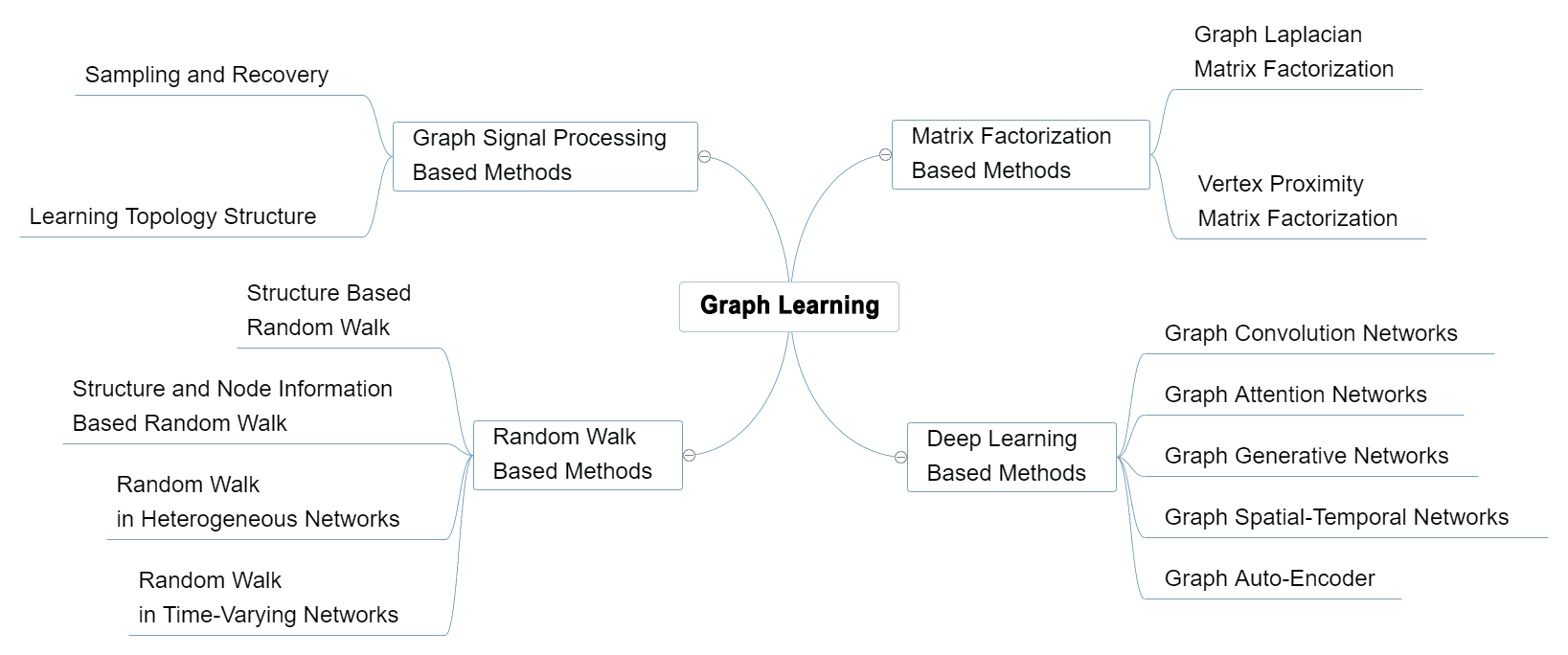}
	\caption{The categorization of graph learning.}\label{fig1}
\end{figure*}

In this survey paper, we try to examine machine learning methods on graphs in a comprehensive manner. As shown in Fig.~\ref{fig1}, we focus on existing methods that fall into the following four categories: graph signal processing (GSP) based methods, matrix factorization based methods, random walk based methods, and deep learning based methods. Roughly speaking, GSP deals with sampling and recovery of graph, and learning topology structure from data. Matrix factorization can be divided into graph Laplacian matrix factorization and vertex proximity matrix factorization. Random walk based methods include structure-based random walk, structure and node information based random walk, random walk in heterogeneous networks, and random walk in time-varying networks. Deep learning based methods include graph convolutional networks, graph attention networks, graph auto-encoder, graph generative networks, and graph spatial-temporal networks. Basically, the model architectures of these methods/techniques differ from each other. This paper presents an extensive review of the state-of-the-art graph learning techniques.

Traditionally, researchers adopt an adjacency matrix to represent a graph, which can only capture the relationship between two adjacent vertices. However, many complex and irregular structures cannot be captured by this simple representation. When we analyze large-scale networks, traditional methods are computationally expensive and hard to be implemented in real-world applications. Therefore, effective representation of these networks is a paramount problem to solve~\cite{zhang2018network}. Network Representation Learning (NRL) proposed in recent years can learn latent features of network vertices with low dimensional representation~\cite{grover2016node2vec,nrl2020access,yu2020cikm}. When the new representation has been learned, previous machine learning methods can be employed for analyzing the graph data as well as discovering relationships hidden in the data.

When complex networks are embedded into a latent, low dimensional space, the structural information and vertex attributes can be preserved~\cite{zhang2018network}. Thus the vertices of networks can be represented by low dimensional vectors. These vectors can be regarded as the features of input in previous machine learning methods. Graph learning methods pave the way for graph analysis in the new representation space, and many graph analytical tasks, such as link prediction, recommendation and classification, can be solved efficiently~\cite{bengio2013representation,guo2020aaai}. Graphical network representation sheds light on various aspects of social life, such as communication patterns, community structure, and information diffusion~\cite{xia2014exploiting,ddcss2020bdr}. According to the attributes of vertices, edges and subgraph, graph learning tasks can be divided into three categories, which are vertices based, edges based, and subgraph based, respectively. The relationships among vertices in a graph can be exploited for, e.g., classification, risk identification, clustering, and community detection~\cite{xia2014community}. By judging the presence of edges between two vertices in graphs, we can perform recommendation and knowledge reasoning, for instance. Based on the classification of subgraphs~\cite{motif2019access}, the graph can be used for, e.g., polymer classification, 3D visual classification, etc. For GSP, it is significant to design suitable graph sampling methods to preserve the features of the original graph, which aims at recovering the original graph efficiently~\cite{leskovec2006sampling}. Graph recovery methods can be used for constructing the original graph in the presence of incomplete data~\cite{wang2015local}. Afterwards, graph learning can be exploited to learn the topology structure from graph data. In summary, graph learning can be used to tackle the following challenges, which are difficult to solve by using traditional graph analysis methods~\cite{zhang2018deep}.

\begin{figure}[]
\centering
\includegraphics[width=0.4\textwidth]{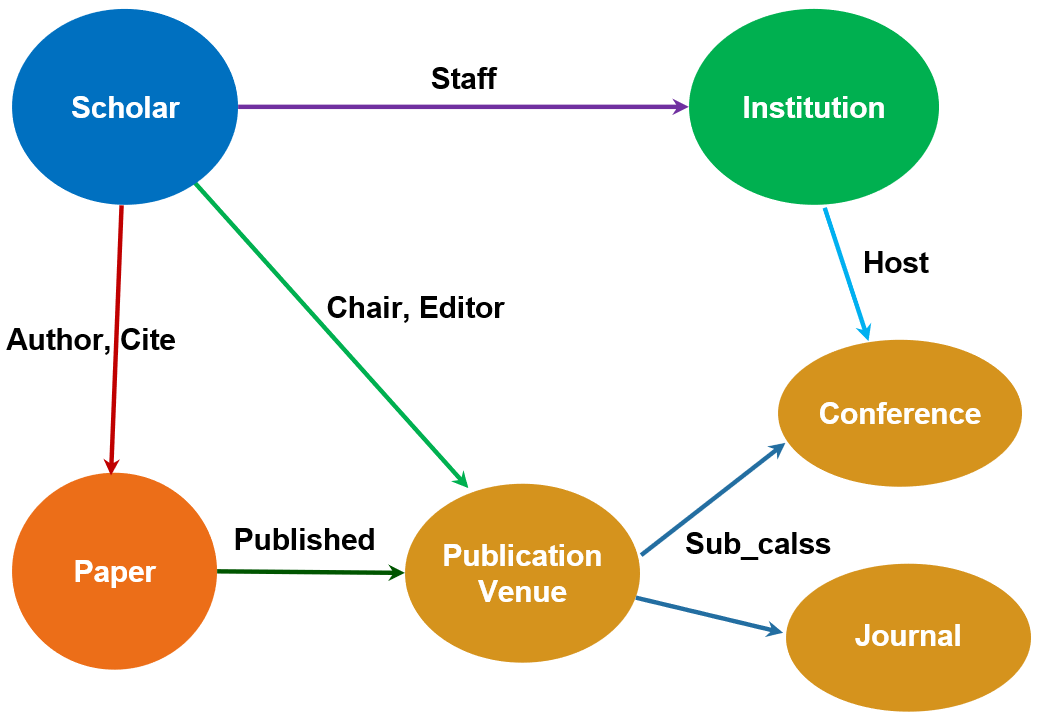}
\caption{Heterogeneous academic network~\cite{xia2017big}.}
\label{fig:Heterogenous}
\end{figure}

\begin{enumerate}
  \item \textbf{Irregular domains:} Data collected by traditional sensors have a clear grid structure. However, graphs lie in an irregular domain (i.e., non-Euclidean space). In contrast to regular domain (i.e., Euclidean space), data in non-Euclidean space are not ordered regularly. Distance is hence difficult to be defined. As a result, basic methods based on traditional machine learning and signal processing cannot be directly generalized to graphs.
  \item \textbf{Heterogeneous networks:} In many cases, networks involved in the traditional graph analysis algorithms are homogeneous. The appropriate modeling methods only consider the direct connection of the network and strip other irrelevant information, which significantly simplifies the processing. However, it is prone to cause information loss. In the real world, the edges among vertices and the types of vertices are usually diverse, such as in the academic network shown in Fig.~\ref{fig:Heterogenous}. Thus it isn't easy to discover potential value from heterogeneous information networks with abundant vertices and edges.
  \item \textbf{Distributed algorithms:} In big social networks, there are often millions of vertices and edges~\cite{xu2020jcdl}. Centralized algorithms cannot handle this since the computational complexity of these algorithms would significantly increase with the growth of vertex number. The design of distributed algorithms for dealing with big networks is a critical problem yet to be solved~\cite{bedru2020big}. One major benefit of distributed algorithms is that the algorithms can be executed in multiple CPUs or GPUs simultaneously, and hence the running time can be reduced significantly.
\end{enumerate}

\subsection{Related Surveys}
There are several surveys that are partially related to the scope of this paper. Unlike these surveys, we aim to provide a comprehensive overview of graph learning methods, with a focus on four specific categories. In particular, graph signal processing is introduced as one approach for graph learning, which is not covered by other surveys.

Goyal and Ferrara~\cite{goyal2018graph} summarized graph embedding methods, such as matrix factorization, random walk and their applications in graph analysis. Cai et al.~\cite{cai2018comprehensive} reviewed graph embedding methods based on problem settings and embedding techniques. Zhang et al.~\cite{zhang2018network} summarized NRL methods based on two categories, i.e., unsupervised NRL and semi-supervised NRL, and discussed their applications. Nickel et al.~\cite{nickel2016review} introduced knowledge extraction methods from two aspects: latent feature models and graph based models. Akoglu et al.~\cite{akoglu2015graph} reviewed state-of-the-art techniques for event detection in data represented as graphs, and their applications in the real world. Zhang et al.~\cite{zhang2018deep} summarized deep learning based methods for graphs, such as graph neural networks (GNNs), graph convolutional networks (GCNs) and graph auto-encoders (GAEs). Wu et al.~\cite{wu2019comprehensive} reviewed state-of-the-art GNN methods and discussed their applications in different fields. Ortega et al.~\cite{ortega2018graph} introduced GSP techniques for representation, sampling and learning, and discussed their applications. Huang et al.~\cite{huang2018graph} examined the applications of GSP in functional brain imaging and addressed the problem of how to perform brain network analysis from signal processing perspective.

In summary, none of the existing surveys provides a comprehensive overview of graph learning. They only cover some parts of graph learning, such as network embedding and deep learning based network representation. The NRL and/or GNN based surveys do not cover the GSP techniques. In contrast, we review GSP techniques in the context of graph learning, as it is an important approach for GNNs. Specifically, this survey paper integrates state-of-the-art machine learning techniques for graph data, gives a general description of graph learning, and discusses its applications in various domains.

\subsection{Contributions and Organization}
The contributions of this paper can be summarized as follows.

\begin{itemize}
    \item\textbf{A comprehensive overview of state-of-the-art graph learning methods:} we present an integral introduction to graph learning methods, including, e.g., technical sketches, application scenarios, and potential research directions.
    \item\textbf{Taxonomy of graph learning:} we give a technical classification of mainstream graph learning methods from the perspective of theoretical models. Technical descriptions are provided wherever appropriate to improve understanding of the taxonomy.
    \item\textbf{Insights into future directions in graph learning:} Besides qualitative analysis of existing methods, we shed light on potential research directions in the field of graph learning through summarizing several open issues and relevant challenges.
\end{itemize}

The rest of this paper is organized as follows. An overview of graph learning approaches containing graph signal processing based methods, matrix factorization based methods, random walk based methods, and deep learning based methods is provided in Section~\ref{sec:algorithm}. The applications of graph learning are examined in Section~\ref{sec:appication}. Some future directions as well as challenges are discussed in Section~\ref{sec:issues}. We conclude the survey in Section~\ref{sec:con}.

\section{Graph Learning Models and Algorithms}\label{sec:algorithm}

The feature vectors that represent various categorical attributes are viewed as the input in previous machine learning methods. However, the mapping from the input feature vectors to the output prediction results need to be handled by graph learning~\cite{goyal2018graph}. Deep learning has been regarded as one of the most successful techniques in artificial intelligence~\cite{lecun2015deep,liu2018artificial}. Extracting complex patterns by exploiting deep learning from a massive amount of irregular data has been found very useful in various fields, such as pattern recognition and image processing. Consequently, how to utilize deep learning techniques to extract patterns from complex graphs has attracted lots of attention. Deep learning on graphs, such as GNNs, GCNs, and GAEs, has been recognized as a powerful technique for graph analysis~\cite{zhang2018deep}. Besides, GSP has also been proposed to deal with graph analysis~\cite{ortega2018graph}. One of the most typical scenarios is that a set of values reside on a set of vertices, and these vertices are connected by edges~\cite{sandryhaila2013discrete}. Graph signals can be adopted to model various phenomena in real world. For example, in social networks, users in Facebook can be viewed as vertices, and their friendships can be modeled as edges. The number of followers of each vertex is marked in this social network. Based on this assumption, many techniques (e.g., convolution, filter, wavelet, etc.) in classical signal processing can be employed for GSP with suitable modifications~\cite{ortega2018graph}.

In this section, we review graph learning models and algorithms under four categories as mentioned before, namely GSP based methods, matrix factorization based methods, random walk based methods, and deep learning based methods. In Table~\ref{tab:abbreviations}, we list the abbreviations used in this paper.

\begin{table}[htbp]
  \centering
  \caption{Definitions of abbreviations}
    \begin{tabular}{cc}
    \toprule
    \textbf{Abbreviation} & \textbf{Definition} \\
    \midrule
    PCA   & Principal component analysis  \\
    NRL   &  Network representation learning \\
    LSTM  &  Long short-term memory (networks)  \\
    GSP   & Graph signal processing \\
    GNN  &  Graph neural network \\
    GMRF  & Gauss markov random field \\
    GCN  & Graph convolutional network \\
    GAT  &  Graph attention network  \\
    GAN   &  Generative adversarial network  \\
    GAE  & Graph auto-encoder \\
    ASP   & Algebraic signal processing \\
     RNN &  Recurrent neural network \\
     CNN & Convolutional neural network \\
    \bottomrule
    \end{tabular}%
  \label{tab:abbreviations}%
\end{table}%

\subsection{Graph Signal Processing}
\begin{figure}[htb]
\centering
\includegraphics[width=0.5\textwidth]{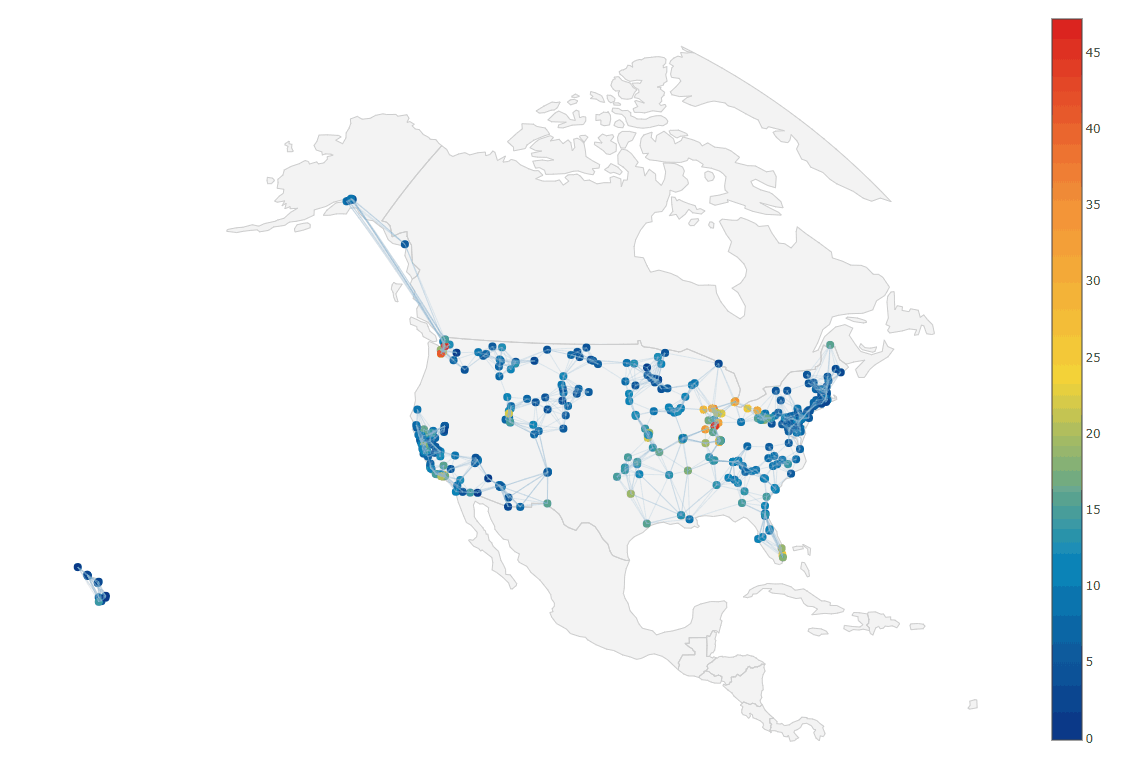}
\caption{The measurements of PM2.5 from different sensors on July 5, 2014 (data source: https://www.epa.gov/).}
\label{fig:US}
\end{figure}

Signal processing is a traditional subject that processes signals defined in regular data domain. In recent years, researchers extend concepts of traditional signal processing into graphs. Classical signal processing techniques and tools such as Fourier transform and filtering can be used to analyze graphs. In general, graphs are a kind of irregular data, which are hard to handle directly. As a complement to learning methods based on structures and models, GSP provides a new perspective of spectral analysis of graphs. Derived from signal processing, GSP can give an explanation of graph property consisting of connectivity, similarity, etc. Fig.~\ref{fig:US} gives a simple example of graph signals at a certain time point, which is defined as observed values. In a graph, the above mentioned observed values can be regarded as graph signals. Each node is then mapped to the real number field in GSP. The main task of GSP is to expand signal processing approaches to mine implicit information in graphs.

\subsubsection{Representation on Graphs}
A meaningful representation of graphs has contributed a lot to the rapid growth of graph learning. There are two main models of GSP, i.e., adjacency matrix based GSP~\cite{sandryhaila2013discrete} and Laplacian based GSP~\cite{shuman2012emerging}. An adjacency matrix based GSP comes from algebraic signal processing (ASP)~\cite{puschel2006algebraic}, which interprets linear signal processing from algebraic theory. Linear signal processing contains signals, filters, signal transformation, etc. It can be applied in both continuous and discrete time domains. The basic assumption of linear algebra is extended to the algebra space in ASP. By selecting signal model appropriately, ASP can obtain different instances in linear signal processing. In adjacency matrix based GSP, the signal model is generated from a shift. Similar to traditional signal processing, a shift in GSP is a filter in graph domain~\cite{sandryhaila2013discrete,sandryhaila2013discrete1,chen2015discrete}. GSP usually defines graph signal models using adjacency matrices as shifts. Signals of a graph are normally defined at vertices.

Laplacian based GSP originates from spectral graph theory. High dimensional data are transferred into a low dimensional space generated by a part of the Laplacian basis~\cite{von2007tutorial}. Some researchers exploited sensor networks~\cite{zhu2012graph} to achieve distributed processing of graph signals. Other researchers solved the problem globally under the assumption that the graph is smooth. Unlike adjacency matrix based GSP, Laplacian matrix is symmetric with real and non-negative edge weights, which is used to index undirected graphs.

Although the models use different matrices as basic shifts, most of the notions in GSP are derived from signal processing. Notions with different definitions in these models may have similar meanings. All of them correspond to concepts in signal processing. Signals in GSP are values defined on graphs, and they are usually written as a vector, \(\bm{s}=[ s_{0} ,s_{1} ,\dots ,s_{N-1}] \in \mathbb{C}^{N}.\) \(N\) is the number of vertices, and each element in the vector represents the value on a vertex. Some studies~\cite{ortega2018graph} allow complex-value signals, even though most applications are based on real-value signals.

In the context of adjacency matrix based GSP, a graph can be represented as a triple \(G(V,E,\bm{W})\), where \(V\) is the vertex set, \(E\) is the edge set and \(\bf{W}\) is the adjacency matrix. With the definition of graphs, we can also define degree matrix \(\bm{D}_{ii}=\bm{d}_{i}\), where \(\bm{D}\) is a diagonal matrix, and \(\bm{d}_{i}\) is the degree of vertex \(i\). Graph Laplacian is defined as \(\bm{L}=\bm{D}-\bm{W}\), and normalized Laplacian is defined as \(\bm{L}_{norm}=\bm{D}^{-1/2}\bm{L}\bm{D}^{-1/2}\). Filters in signal processing can be seen as a function that amplifies or reduces relevant frequencies, eliminating irrelevant ones. Matrix multiplication in linear space equals to scale changing, which is identical with filter operation in frequency domain. It is obvious that we can use matrix multiplication as a filter in GSP, which is written as \(\bm{s}_{out} =\bm{Hs}_{in}\), where $\bm{H}$ stands for a filter.

Shift is an important concept to describe variation in signal, and time-invariant signals are used frequently~\cite{sandryhaila2013discrete}. In fact, there are different choices of shifts in GSP. Adjacency matrix based GSP uses \(\bm{A}\) as shift. Laplacian based GSP uses \(\bf{L}\)~\cite{shuman2012emerging}, and some researchers also use other matrices \cite{gavili2017shift}. By following time invariance in traditional signal processing, shift invariance is defined in GSP. If filters are commutative with shift, they are shift-invariant, which can be written as \(\bm{AH}=\bm{HA}\). It is proved that shift-invariant filter can be represented by the shift. The properties of shift are vital, and they determine the fashion of other definitions such as Fourier transform and frequency.

\begin{figure}[htb]
\centering
\subfigure[The maximum  frequency]{
\label{maxfre}
\includegraphics[width=0.5\textwidth]{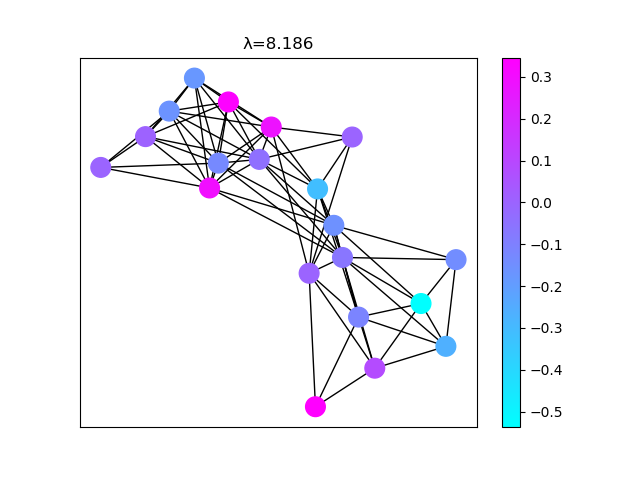}}
\subfigure[The minimum frequency]{
\label{minfre}
\includegraphics[width=0.5\textwidth]{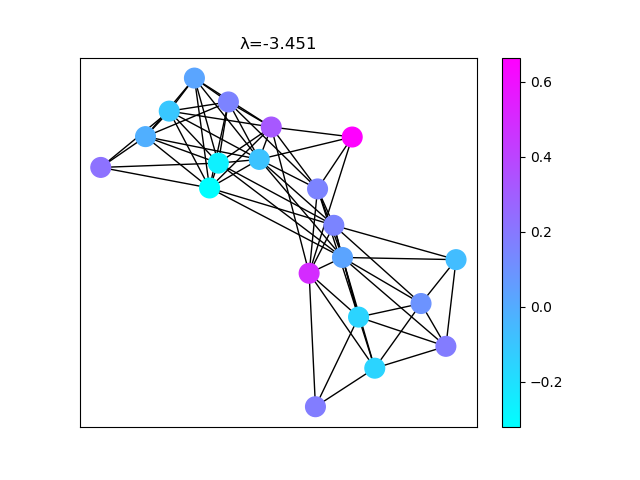}}
\caption{Illustration of difference between minimum and maximum frequencies.}
\label{fig:lambda}
\end{figure}

In adjacency matrix based GSP, eigenvalue decomposition of shift \(\bm{A}\) is \(\bm{A}=\bm{V\Lambda V}^{-1}\). \(\bm{V}\) is the matrix of eigenvectors \([\bm{v}_{0}, \bm{v}_{1},\dots,\bm{v}_{N-1}]\) and
\[
  \bm{\Lambda} =
  \begin{bmatrix}
    \bm{\lambda}_{0} & & \\
    & \ddots & \\
    & & \bm{\lambda}_{N-1}
  \end{bmatrix}
\]
is a diagonal matrix of eigenvalues. The Fourier transform matrix is the inverse of \(\bm{V}\), i.e., \(\bm{F}=\bm{V}^{-1}\). Frequency of shift is defined as total variation, which states the difference after shift \[TV_{G} =||\bm{v}_{k} -\frac{1}{\lambda _{max}} \bm{A}\bm{v}_{k} ||_{1},\] where \(\frac{1}{\lambda _{max}}\) is a normalized factor of matrix. It means that the frequencies of eigenvalue far away from the largest eigenvalues on complex plane are large. A large frequency means that signals are changed with a large scale after shift filtering. The differences between minimum and maximum \(\bm{\lambda}\) can be seen in Fig.~\ref{fig:lambda}. Generally, the total variation tends to be relatively low with larger frequency, and vice versa. Eigenvectors of larger eigenvalues can be used to construct low-frequency filters, which capture fundamental characteristics, and smaller ones can be employed to capture the variation among neighbor nodes.

For topology learning problems, we can distinguish the corresponding solutions depending on known information. When topology information is partly known, we can use the known information to infer the whole graph. In case the topology information is unknown while we still can observe the signals on the graph, the topology structure has to be inferred from the signals. The former one is often solved as a sampling and recovery problem, and blind topology inference is also known as graph topology (or structure) learning.

\subsubsection{Sampling and Recovery}
Sampling is not a new concept defined in GSP. In conventional signal processing, we normally need to reconstruct original signals with the least samples and retain all information of original signals for a sampling problem. Few samples lead to the lack of information and more samples need more space to store. The well-known Nyquist-Shannon sampling theorem gives the sufficient condition of perfect recovery of signals in time domain.

Researchers have migrated the sampling theories into GSP to study the sampling problem on graphs. As the volume of data is large in some real-world applications such as sensor networks and social networks, sampling less and recovering better are vital for GSP. In fact, most algorithms and frameworks solving sampling problems require that the graph models correlations within signals observed on it~\cite{pasdeloup2015graph}. The sampling problem can be defined as reconstructing signals from samples on a subset of vertices, and signals in it are usually band-limited. Nyquist-Shannon sampling theorem was extended to graph signals in~\cite{anis2014towards}. Based on the normalized Laplacian matrix, sampling theorem and cut-off frequency are defined for GSP. Moreover, the authors provided a method for computing cut-off frequency from a given sampling set and a method for choosing sampling set for a given bandwidth. It should be noted that the sampling theorem proposed therein is merely applied to undirected graph. As Laplacian matrix represents undirected graphs only, sampling theory for directed graph adopts adjacent matrix. An optimal operator with a guarantee for perfect recovery was proposed in~\cite{chen2015discrete}, and it is robust to noise for general graphs.

One of the explicit distinctions between classical signal processing and GSP is that signals of the former fall in regular domain while the latter falls in irregular domain. For sampling and recovery problems, classical signal processing samples successive signals and can recover successive signals from samplings. GSP samples a discrete sequence, and recovers the original sequences from samplings. By following this order, the solution is generally separated into two parts, i.e., finding sampling vertex sets and reconstructing original signals based on various models.

When the size of the dataset is small, we can handle the signal and shift directly. However, for a large-scale dataset, some algorithms require matrix decomposition to obtain frequencies and save eigenvalues in the procedure, which are almost impossible to realize. As a simple technique applicable to large-scale datasets, a random method can also be used in sampling. Puy et al.~\cite{puy2018random} proposed two sample strategies: a non-adaptive one depending on a parameter and an adaptive random sampling strategy. By relaxing the optimized constraint, they extended random sampling to large scale graphs. Another common strategy is greedy sampling. For example, Shomorony and Avestimehr~\cite{shomorony2014sampling} proposed an efficient method based on linear algebraic conditions that can exactly compute cut-off frequency. Chamon and Ribeiro~\cite{chamon2018greedy} provided near-optimal guarantee for greedy sampling, which guarantees the performance of greedy sampling in the worst cases.

All of the sampling strategies mentioned above can be categorized as selecting sampling, where signals are observed on a subset of vertices \cite{chamon2018greedy}. Besides selecting sampling, there exists a type of sampling called aggregation sampling~\cite{marques2016sampling}, which uses observations taken at a single vertex as input, containing a sequential applications of graph shift operator.

Similar to classical signal processing, the reconstruction task on graphs can also be interpreted as data interpolation problem \cite{narang2013signal}. By projecting the samples on a proper signal space, researchers obtain interpolated signals. Least squares reconstruction is an available method in practice. Gadde and Ortega~\cite{gadde2015probabilistic} defined a generative model for signal recovery derived from a pairwise Gaussian random field (GRF) and a covariance matrix on graphs. Under sampling theorem, the reconstruction of graph signals can be viewed as the maximum posterior inference of GRF with low-rank approximation. Wang et al.~\cite{wang2015distributed} aimed at achieving the distributed reconstruction of time-varying band limited signal, where the distributed least squares reconstruction (DLSR) was proposed to recover the signals iteratively. DLSR can track time-varying signals and achieve perfect reconstruction. Di Lorenzo et al.~\cite{di2016adaptive} proposed a linear mean squares (LMS) strategy for adaptive estimation. LMS enables online reconstruction and tracking from the observation on a subset of vertices. It also allows the subset to vary over time. Moreover, a sparse online estimation was proposed to solve the problems with unknown bandwidth.

Another common technique for recovering original signals is smoothness. Smoothness is used for inferring missing values in graph signals with low frequencies. Wang et al.~\cite{wang2015local} defined the concept of local set. Based on the definition of graph signals, two iterative methods were proposed to recover band limited signals on graphs. Besides, Romero et al.~\cite{romero2017kernel} advocated kernel regression as a framework for GSP modeling and reconstruction. For parameter selection in estimators, two multi-kernel methods were proposed to solve a single optimization problem as well. In addition, some researchers investigated different recovery problems with compressed sensing~\cite{nagahara2015discrete}.

In addition, there exists some researches on sampling of different kinds of signals such as smooth graph signals, piece-wise constant signals and piece-wise smooth signals~\cite{chen2015signal}. Chen et al. ~\cite{chen2015signal} gave a uniform framework to analyze graph signals. The reconstruction of a known graph signal was studied in~\cite{segarra2016reconstruction}, where the signal is sparse, which means only a few vertices are non-zeros. Three kinds of reconstruction schemes corresponding to various seeding patterns were examined. By analyzing single simultaneous injection, single successive value injection, and multiple successive simultaneous injections, the conditions for perfect reconstruction on any vertices were derived.

\subsubsection{Learning Topology Structure from Data}
In most application scenes, graphs are constructed according to connections of entity correlations. For example, in sensor networks, the correlations between sensors are often consistent with geographic distance. Edges in social networks are defined as relations such as friends or colleagues~\cite{tn2020sigweb}. In biochemical networks, edges are generated by interactions. Although GSP is an efficient framework for solving problems on graphs such as sampling, reconstruction, and detection, there lacks a step to extract relations from datasets. Connections exist in many datasets without explicit records. Fortunately, they can be inferred in many ways.

As a result, researchers want to learn complete graphs from datasets. The problem of learning graph from a dataset is stated as estimating graph Laplacian, or graph topology~\cite{egilmez2017graph}. Generally, they require the graph to satisfy some properties, such as sparsity and smoothness. Smoothness is a widespread assumption in networks generated from datasets. Therefore, it is usually used to constrain observed signals and provide a rational guarantee for graph signals. Researchers have applied it to graph topology learning. The intuition behind smoothness based algorithms is that most signals on graph are stationary, and the result filtered by shift tends to be the lowest frequency. Dong et al.~\cite{dong2016learning} adopted a factor analysis model for graph signals, and also imposed a Gaussian prior on latent variables to obtain a Principal Component Analysis (PCA) like representation. Kalofolias~\cite{kalofolias2016learn} formulated the objective as a weighted \(l_1\) problem and designed a general framework to solve it.

Gauss Markov Random Field (GMRF) is also a widely used theory for graph topology learning in GSP \cite{egilmez2017graph,pavez2016generalized,pavez2018learning}. The models of GRMF based graph topology learning select graphs that are more likely to generate signals which are similar to the ones generated by GMRF. Egilmez et al.~\cite{egilmez2017graph} formulated the problem as a maximum posterior parameter estimation of GMRF, and the graph Laplacian is a precision matrix. Pavez and Ortega~\cite{pavez2016generalized} also formulated the problem as a precision matrix estimation, and the rows and columns are updated iteratively by optimizing a quadratic problem. Both of them restrict the result matrix, which should be Laplacian. In~\cite{pavez2018learning}, Pavez et al. chose a two steps framework to find the structure of the underlying graph. First, a graph topology inference step is employed to select a proper topology. Then, a generalized graph Laplacian is estimated. An error bound of Laplacian estimation is computed. In the next step, the error bound can be utilized to obtain a matrix in a specific form as the precision matrix estimation. It is one of the first work that suggests adjusting the model to obtain a graph satisfying the requirement of various problems.

Diffusion is also a relevant model that can be exploited to solve the topology interfering problem~\cite{pasdeloup2015graph,pasdeloup2018characterization,segarra2017network,thanou2017learning}. Diffusion refers to that the node continuously influences its neighborhoods. In graphs, nodes with larger values will have higher influence on their neighborhood nodes. Using a few components to represent signals will help to find the main factors of signal formation. The models of diffusion are often under the assumption of independent identically-distributed signals. Pasdeloup et al.~\cite{pasdeloup2018characterization} gave the concept of valid graphs to explain signals and assumed that the signals are observed after diffusion. Segarra et al.~\cite{segarra2017network} agreed that there exists a diffusion process in the shift, and the signals can be observed. The signals in~\cite{thanou2017learning} were explained as a linear combination of a few components.

For time series recorded in data, researchers tried to construct time-sequential networks. For instance, Mei and Moura~\cite{mei2016signal} proposed a methodology to estimate graphs, which considers both time and space dependencies and models them by auto-regressive process. Segarra et al.~\cite{segarra2017blind} proposed a method that can be seen as an extension of graph learning. The aim of the paper was to solve the problem of joint identification of a graph filter and its input signal.

For recovery methods, a well-known partial inference problem is recommendation~\cite{narang2013signal, xia2013recAccess, huang2018rating}. The typical algorithm used in recommendation is collaborative filtering (CF)~\cite{xia2016scientific}. Given the observed ratings in a matrix, the objective of CF is to estimate the full rating matrix. Huang et al.~\cite{huang2018rating} demonstrated that collaborative filtering could be viewed as a specific band-stop graph filter on networks representing correlations between users and items. Furthermore, linear latent factor methods can also be modeled as band limited interpretation problem.

\subsubsection{Discussion}
GSP algorithms have strict limitations on experimental data, thus leading to less real-world applications. Moreover, GSP algorithms require the input data to be exactly the whole graph, which means that part of graph data cannot be the input. Therefore, the computational complexity of this kind of methods could be significantly high. In comparison with other kinds of graph learning methods, the scalability of GSP algorithms is relatively poor.

\subsection{Matrix Factorization Based Methods}
Matrix factorization is a method of simplifying a matrix into its components. These components have a lower dimension and could be used to represent the original information of a network, such as relationships among nodes. Matrix factorization based graph learning methods adopt a matrix to represent graph characteristics like vertex pairwise similarity, and the vertex embedding can be achieved by factorizing this matrix~\cite{he2004locality}. Early graph learning approaches usually utilized matrix factorization based methods to solve the graph embedding problem. The input of matrix factorization is the non-relational high dimensional data feature represented as a graph. The output of matrix factorization is a set of vertex embedding. If the input data lies in a low dimensional manifold, the graph learning for embedding can be treated as a dimension-reduced problem that preserves the structure information. There are mainly two types of matrix factorization based graph learning. One is graph Laplacian matrix factorization, and the other is vertex proximity matrix factorization.

\subsubsection{Graph Laplacian Matrix Factorization}
The preserved graph characteristics can be expressed as pairwise vertex similarities. Generally, there are two kinds of graph Laplacian matrix factorization, i.e., transductive and inductive matrix factorization. The former only embeds the vertices contained in the training set, and the latter can embed the vertices that are not contained in the training set. The general framework has been designed in~\cite{chen2014unified}, and the graph Laplacian matrix factorization based graph learning methods have been summarized in~\cite{yan2007graph}. The Euclidean distance between two feature vectors is directly adopted in the initial Metric Multidimensional Scaling (MDS)~\cite{borg2003modern} to find the optimal embedding. The neighborhoods of vertices are not considered in the MDS, i.e., any pair of training instances are considered as connected. The data feature is extracted by constructing a $k$ nearest neighbor graph, and the subsequent studies \cite{he2004locality,balasubramanian2002isomap,anderson1985eigenvalues,roweis2000nonlinear} tackle this issue. The top $k$ similar neighbors of each vertex are connected with itself. A similar matrix is calculated by exploiting different methods, and thus the graph characteristics can be preserved as much as possible.

Recently, researchers have designed more sophisticated models. The performance of earlier matrix factorization model Locality Preserving Projection (LPP) can be improved by introducing an anchor taking advantage of Anchorgraph-based Locality Preserving Projection (AgLPP)~\cite{jiang2016dimensionality,wan2019your}. The graph structure can be captured by using a local regression model and a global regression process based on Local and Global Regressive Mapping (LGRM)~\cite{yang2010local}. The global geometry can be preserved by using local spline regression~\cite{xiang2008nonlinear}.

More information can be preserved by exploiting the auxiliary information. An adjacency graph and a labelled graph were constructed in~\cite{cai2007spectral}. The objective function of LPP preserves the local geometric structure of the datasets~\cite{he2004locality}. An adjacency graph and a relational feedback graph were constructed in~\cite{he2004learning} as well. The graph Laplacian regularization, k-means and PCA were considered in RF-Semi-NMF-PCA simultaneously~\cite{allab2017a}. Other works, e.g.,~\cite{vandenberghe1996semidefinite}, adopt semi-definite programming to learn the adjacency graph that maximizes the pairwise distances.

\subsubsection{Vertex Proximity Matrix Factorization}
Apart from solving the above generalized eigenvalue problem, another approach of matrix factorization is to factorize vertex  proximity matrix directly. In general, matrix factorization can be used to learn the graph structure from non-relational data, and it is applicable to learn homogeneous graphs.

Based on matrix factorization, vertex proximity can be approximated in a low dimensional space. The objective of preserving vertex proximity is to minimize the error. The Singular Value Decomposition (SVD) of vertex proximity matrix was adopted in~\cite{golub1970singular}. There are some other approaches such as regularized Gaussian matrix factorization~\cite{ahmed2013distributed}, low-rank matrix factorization~\cite{yang2015network}, for solving SVD.

\subsubsection{Discussion}
Matrix factorization algorithms operate on an interaction matrix to decompose several lower dimension matrices. The process brings some drawbacks. For example, the algorithms require a large memory when the decomposed matrices become large. In addition, matrix factorization algorithms are not applicable to supervised or semi-supervised tasks with the training process.

\subsection{Random Walk Based Methods}
\begin{figure}[htb]
\centering
\subfigure[Image in Euclidean space]{
\label{enc}
\includegraphics[width=0.4\textwidth]{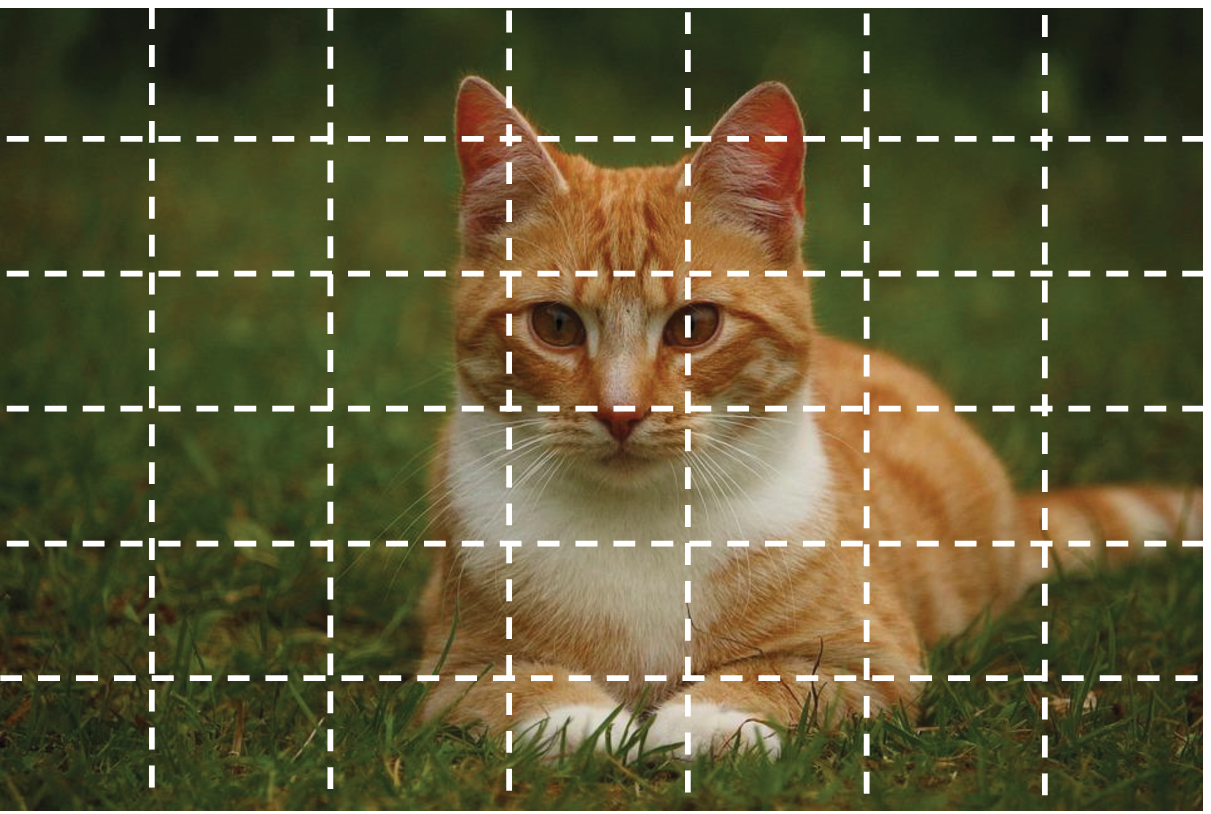}}
\subfigure[Graph in non-Euclidean space]{
\label{noenc}
\includegraphics[width=0.4\textwidth]{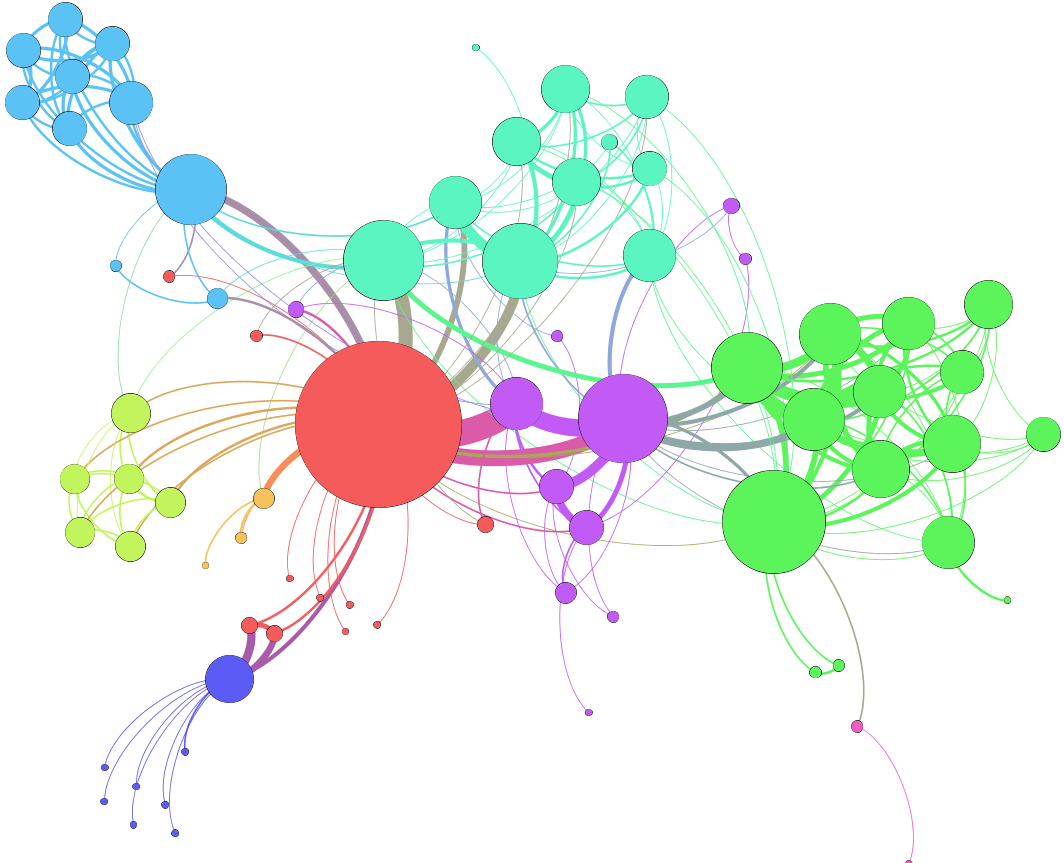}}
\caption{An example of NRL mapping an image from Euclidean space into non-Euclidean space.}
\label{fig:NRL}
\end{figure}

Random walk is a convenient and effective way to sample networks~\cite{xia2019random,xia2014mvcwalker}. This method can generate sequences of nodes meanwhile preserving original relations between nodes. Based on network structure, NRL can generate feature vectors of vertices so that downstream tasks can mine network information in a low dimensional space. An example of NRL is shown in Fig.~\ref{fig:NRL}. The image in Euclidean space is shown in Fig.~\ref{enc}, and the corresponding graph in non-Euclidean space is shown in Fig.~\ref{noenc}. As one of the most successful NRL algorithms, random walks play an important role in dimensionality reduction.

\subsubsection{Structure Based Random Walks}
Graph-structured data have various data types and structures. The information encoded in a graph is related to graph structure and vertex attributes, which are the two key factors affecting the reasoning of networks. In real-world applications, many networks only have structural information, but lack vertex attribute information. How to identify network structure information effectively, such as important vertices and invisible links, attracts the interest of network scientists~\cite{al2018analysis}. Graph data have high dimensional characteristics. Traditional network analysis methods cannot be used for analyzing graph data in a continuous space.

In recent years, various NRL methods have been proposed, which preserve rich structural information of networks. DeepWalk~\cite{perozzi2014deepwalk} and Node2vec~\cite{grover2016node2vec} are two representative methods for generating network representation of basic network topology information. These methods use random walk models to generate random sequences on networks. By treating the vertices as words and the generated random sequences of vertices as word sequences (sentences), the models can learn the embedding representation of the vertices by inputting these sequences into the Word2vec model~\cite{levy2014neural,rong2014word2vec,goldberg2014word2vec}. The principle of the learning model is to maximize the co-occurrence probability of vertices such as Word2vec. In addition, Node2vec shows that network has complex structural characteristics, and different network structure samplings can obtain different results. The sampling mode of DeepWalk is not enough to capture the diversity of connection patterns in networks. Node2vec designs a random walk sampling strategy, which can sample the networks with the preference of breadth-first sampling and depth-first sampling by adjusting the parameters.

The NRL algorithms mentioned above focused on the first-order proximity information of vertices. Tang et al.~\cite{tang2015www} proposed a method called LINE for large-scale network embedding. LINE can maintain the first and second order approximations. The first-order neighbor refers to the one-hop neighbor between two vertices, and the second-order neighbor is the neighbor with two hops. LINE is not a deep learning based model, but it is often compared with these edge modeling based methods.

It has been proved that the network structure information plays an important role in various network analysis tasks. In addition to this structural information, network attributes in the original network space are also critical in modeling the formation and evolution of the network~\cite{wang2019sustainable}.

\subsubsection{Structure and Vertex Information Based Random Walks}
In addition to network topology, many types of networks also have rich vertex information, such as vertex content or label in networks. Yang et al.~\cite{yang2015network} proposed an algorithm called TADW. The model is based on DeepWalk and considers the text information of vertices. The MMDW~\cite{tu2016max} is another model based on DeepWalk, which is a kind of semi-supervised network embedding algorithm, by leveraging labelling
information of vertices to enhance the performance.

Focusing on the structural identity of nodes, Ribeiro et al.~\cite{ribeiro2017struc2vec} formulated a framework named Struc2vec. The framework considers nodes with similar local structure rather than neighborhood and labels of nodes. With hierarchy to evaluate structural similarity, the framework constrains structural similarity more stringently. The experiments indicate that DeepWalk and Node2vec are worse than Struc2vec which considers structural identity. There are some other NRL models, such as Planetoid~\cite{yang2016revisiting}, which learn network representation using the feature of network structure and vertex attribute information. It is well known that vertex attributes provide effective information for improving network representation and help to learn embedded vector space. In the case of relatively sparse network topology, vertex attribute information can be used as supplementary information to improve the accuracy of representation. In practice, how to use vertex information effectively and how to apply this information to network vertex embedding are the main challenges in NRL.

Researchers not only investigate random walk based NRL on vertices but also on graphs. Adhikari et al.~\cite{adhikari2017distributed} proposed an unsupervised scalable algorithm, Sub2Vec, to learn arbitrary subgraph. To be more specific, they proposed a method to measure the similarities between subgraphs without disturbing local proximity. Narayanan et al.~\cite{narayanan2017graph2vec} proposed graph2vec, which is a neural embedding framework. Modeling on neural document embedding models, graph2vec takes a graph as a document and the subgraph around words as vertices. By migrating the model to graphs, the performance of graph2vec significantly exceeds other substructure representation learning algorithms.

Generally, random walk can be regarded as a Markov process. The next state of the process is only related to last state, which is known as Markov chain. Inspired by vertex-reinforced random walks, Benson et al.~\cite{benson2017spacey} presented spacey random walk, a non-Markovian stochastic process. As a specific type of a more general class of vertex-reinforced random walks, it takes the view that the probability of time remained on each vertex relates to the long term behavior of dynamical systems. They proved that dynamical systems can converge to a stationary distribution under sufficient conditions.

Recently, with the development of Generative Adversarial Network (GAN), researchers combined random walks with the GAN method~\cite{wang2018graphgan, bojchevski2018netgan}. Existing research on NRL can be divided into generative models and discriminative models. GraphGAN~\cite{wang2018graphgan} integrated these two kinds of models and played a game-theoretical minimax game. With the process of the game, the performance of the two models can be strengthened. Random walk is used as a generator in the game. NetGAN~\cite{bojchevski2018netgan} is a generative model that can model network in real applications. The method takes the distribution of biased random walk as input, and can produce graphs with known patterns. It preserves important topology properties and does not need to define them in model definition.

\subsubsection{Random Walks in Heterogeneous Networks}
In reality, most networks contain more than one type of vertex, and hence networks are heterogeneous. Different from homogeneous NRL, heterogenous NRL should well reserve various relationships among different vertices~\cite{shi2017survey}. Considering the ubiquitous existence of heterogeneous networks, many efforts have been made to learn network representations of heterogeneous networks. Compared to homogeneous NRL, the proximity among entities in heterogeneous NRL is more than a simple measure of distance or closeness. The semantics among vertices and links should be considered. Some typical scenarios include knowledge graphs and social networks.

Knowledge graph is a popular research domain in recent years. A vital part in knowledge base population is relational inference. The central problem of relational inference is inferring unknown knowledge from the existing facts in knowledge bases~\cite{lao2010relational}. There are three types of common relational inference method in general: statistical relational learning (SRL), latent factor models (LFM) and random walk models (RWM). Relational learning methods based on statistics lack generality and scalability. As a result, latent factor model based graph embedding and relational paths based random walk have been adopted more widely.

In a knowledge graph, there exist various vertices and various types of relationships among different vertices. For example, in a scholar related knowledge graph~\cite{xia2017big,liu2020wos}, the types of vertices include scholar, paper, publication venue, institution, etc. The types of relationships include coauthor, citation, publication, etc. The key idea of knowledge graph embedding is to embed vertices and their relationships into a low dimensional vector space, while the inherent structure of the knowledge graph can be reserved~\cite{wang2017knowledge}.

For relational paths based random walk, the path ranking algorithm (PRA) is a path finding method using random walks to generate relational features on graph data~\cite{lao2011random}. Random walks in PRA are with restart, and combine features with a logistic regression. However, PRA cannot predict connection between two vertices if there does not exist a path between them. Gardner et al.~\cite{gardner2013improving,gardner2014incorporating} introduced two ways to improve the performance of PRA. One method enables more efficient processing to incorporate new corpus into knowledge base, while the other method uses vector space to reduce the sparsity of surface forms. To resolve cascade errors in knowledge construction, Wang and Cohen~\cite{wang2015joint} proposed a joint information extraction and knowledge base based model with a recursive random walk. Using latent context of the text, the model obtains additional improvement. Liu et al.~\cite{liu2016hierarchical} developed a new random walk based learning algorithm named Hierarchical Random-walk inference (HiRi). It is a two-tier scheme: the upper tier recognizes relational sequence pattern, and the lower tier captures information from subgraphs of knowledge bases.

Another widely-investigated type of heterogeneous networks is social networks, such as online social networks and location based social networks. Social networks are heterogeneous in nature because of the different types of vertices and relations. There are two main ways to embed heterogeneous social networks, including meta path-based approaches and random walk-based approaches.

A meta path in heterogeneous networks is defined as a sequence of vertex types encoding significant composite relations among various types of vertices. Aiming to employ the rich information in social networks by exploiting various types of relationships among vertices, Fu et al.~\cite{fu2017hin2vec} proposed HIN2Vec, which is a representation learning framework based on meta-paths. HIN2Vec is a neural network model and the meta-paths are well embedded based on two independent phases, i.e., training data preparation and representation learning. Experimental results on various social network datasets show that HIN2Vec model is able to automatically learn vertex vector in heterogeneous networks to support a variety of applications. Metapath2vec~\cite{dong2017metapath2vec} was designed by formalizing meta-path based random walks to construct the neighborhoods of a vertex in heterogeneous networks. It takes the advantage of a heterogeneous skip-gram model to perform vertex embedding.

Meta path based methods require either prior knowledge for optimal meta-path selection or extended computations for path length selection. To overcome these challenges, random walk based approaches have been proposed. Hussein et al.~\cite{hussein2018meta} proposed the JUST model, which is a heterogeneous graph embedding approach using random walks with jump and stay strategies so that the aforementioned bias can be overcomed effectively. Another method  which does not require prior knowledge for meta-path definition is MPDRL~\cite{wanreinforcement}, meta-path discovery with reinforcement earning. This method employs the reinforcement learning algorithm to perform multi-hop reasoning to generate path instances and then further summarizes the important meta-paths using the Lowest Common Ancestor principle. Shi et al.~\cite{shi2019heterogeneous} proposed the HERec model, which utilizes the heterogeneous information network embedding for providing accurate recommendations in social networks. HERec is designed based on a random walk based approach for generating meaningful vertex sequences for heterogeneous network embedding. HERec can effectively adopt the auxiliary information in heterogeneous information networks. Other typical heterogeneous social network embedding approaches include, e.g., PTE~\cite{tang2015pte} and SHNE~\cite{zhang2019shne}.

\subsubsection{Random Walks in Time-varying Networks}

Network is evolving over time, which means that new vertices may emerge and new relations may appear. Therefore, it is significant to capture the temporal behaviour of networks in network analysis. Many efforts have been made to learn time-varying network embedding (e.g., dynamic networks or temporal networks)~\cite{netembedding2020csr}. In contrast to static network embedding, time-varying NRL should consider the network dynamics, which means that old relationships may become invalid and new links may appear.

The key of time-varying NRL is to find a suitable way to incorporate the time characteristic into embedding via reasonable updating approaches. Nguyen et al.~\cite{nguyen2018continuous} proposed the CTDNE model for continuous dynamic network embedding based on random walk with "chronological" paths which can only move forward as time goes on. Their model is more suitable for time-dependent network representation that can capture the important temporal characteristics of continuous-time dynamic networks. Results on various datasets show that CTDNE outperforms static NRL approaches. Zuo et al.~\cite{zuo2018embedding} proposed the HTNE model which is a temporal NRL approach based on the Hawkes process. HTNE can well integrate the Hawkes process into network embedding so that the influence of historical neighbors on the current neighbors can be accurately captured.


For unseen vertices in a dynamical network, GraphSAGE~\cite{hamilton2017inductive} was presented to efficiently generate embeddings for new vertices in network. In contrast to methods that training embedding for every vertex in the network, GraphSAGE designs a function to generate embedding for a vertex with features of the neighborhoods locally. After sampling neighbors of a vertex, GraphSAGE uses different aggregators to update the embedding of the vertex. However, current graph neural methods are proficient of only learning local neighborhood information and cannot directly explore the higher-order proximity and the community structure of graphs.

\subsubsection{Discussion}
As mentioned before, random walk is a fundamental way to sample networks. The sequences of nodes could preserve the information of network structure. However, there are some disadvantages of this method. For example, random walk relies on random strategies, which creates some uncertain relations of nodes. To reduce this uncertainty, it needs to increase the number of samples, which will significantly increase the complexity of algorithms. Some random walk variants could preserve local and global information of networks, but they might not be effective in adjusting parameters to adapt to different types of networks.

\subsection{Deep Learning on Graphs}
Deep learning is one of the hottest areas over the past few years. Nevertheless, it is an attractive and challenging task to extend the existing neural network models, such as Recurrent Neural Networks (RNNs) or Convolutional Neural Networks (CNNs), to graph data. Gori et al.~\cite{gori2005new} proposed a GNN model based on recursive neural network. In this model, a transfer function is implemented, which maps the graph or its vertices to an m-dimensional Euclidean space. In recent years, lots of GNN models have been proposed.
\subsubsection{Graph Convolutional Networks}
GCN works on the basis of grid structure domain and graph structure domain~\cite{bruna2013spectral}.

\textbf{Time Domain and Spectral Methods}. Convolution is one of a common operation in deep learning. However, since graph lacks a grid structure, standard convolution over images or text cannot be directly applied to graphs. Bruna et al.~\cite{bruna2013spectral} extended the CNN algorithm from image processing to the graph using the graph Laplacian matrix, dubbed as spectral graph CNN. The main idea is similar to Fourier basis for signal processing. Based on~\cite{bruna2013spectral}, Henaff et al.~\cite{henaff2015deep} defined kernels to reduced the learning parameters by analogizing the local connection of CNNs on the image. Defferrard et al.~\cite{defferrard2016convolutional} provided two ways for generalizing CNNs to graph structure data based on graph theory. One method is to reduce the parameters by using polynomial kernel, and this method can be accelerated by using Chebyshev polynomial approximation. The other method is the special pooling method, which is pooling on the binary tree constructed from vertices. An improved version of \cite{defferrard2016convolutional} was introduced by Kipf and Welling~\cite{kipf2016semi}. The proposed method is a semi-supervised learning method for graphs. The algorithm employs an excellent and straightforward neural network followed by a layer-by-layer propagation rule, which is based on the first-order approximation of spectral convolution on the graph and can be directly acted on the graph.

There are some other time domain based methods. Based on the mixture model of CNNs, for instance, Monti et al.~\cite{monti2017geometric} generalized the CNN to non-Euclidean space. Zhou and Li~\cite{zhou2017graph} proposed a new CNN graph modeling framework, which designs two modules for graph structure data: K-order convolution operator and adaptive filtering module. In addition, the high-order adaptive graph convolution network (HA-GCN) framework proposed in~\cite{zhou2017graph} is a general architecture that is suitable for many applications of vertices and graph centers. Manessi et al.~\cite{manessi2017dynamic} proposed a dynamic graph convolution network algorithm for dynamic graphs. The core idea of the algorithm is to combine the expansion of graph convolution with the improved Long Short Term-Memory networks (LSTM) algorithm, and then train and learn the downstream recursive unit by using graph structure data and vertex features. The spectral based NRL methods have many applications, such as vertex classification~\cite{kipf2016semi}, traffic forecasting~\cite{li2017diffusion,yu2017spatio}, and action recognition~\cite{yan2018spatial}.

\textbf{Space Domain and Spatial Methods}. Spectral graph theory provides a convolution method on graphs, but many NRL methods directly use convolution operation on graphs in space domain. Niepert et al.~\cite{niepert2016learning} applied graph labeling procedures such as Weisfeiler-Lehman kernel on graphs to generate unique order of vertices. The generated sub-graphs can be fed to the traditional CNN operation in space domain. Duvenaud et al.~\cite{duvenaud2015convolutional} designed Neural fingerprints (FP), which is a spatial method using the first-order neighbors similar to the GCN algorithm. Atwood and Towsley~\cite{atwood2016diffusion} proposed another convolution method, called diffusion-convolutional neural network, which incorporates transfer probability matrix and replaces the characteristic basis of convolution with diffusion basis. Gilmer et al.~\cite{gilmer2017neural} reformulated existing models into a single common framework, and exploited this framework to discover new variations. Allamanis et al.~\cite{allamanis2017learning} represented the structure of code from syntactic and semantic, and utilized the GNN method to recognize program structures.

Zhuang and Ma~\cite{zhuang2018dual} designed dual graph convolution networks (DGCN), which use diffusion basis and adjacency basis. DGCN uses two convolutions: one is the characteristic form of polynomial filter, and the other is to replace the adjacency matrix with the PPMI (Positive Pointwise Mutual Information) of the transition probability~\cite{levy2014neural}. Dai et al.~\cite{dai2018learning} proposed the SSE algorithm, which uses asynchronous random to learn vertex representation so as to improve learning efficiency. In this model, a recursive method is adopted to learn vertex latent representation and the sampled batch data are utilized to update parameters. The recursive function of SSE is calculated from the weighted average of historical state and new state. Zhu et al.~\cite{aaai20-zhu} proposed a graph smoothing splines neural network which exploits non-smoothing node features and global topological knowledge such as centrality for graph classification. Gao et al.~\cite{gao2018large} proposed a large scale graph convolution network (LGCN) based on vertex feature information. In order to adapt to the scene of large scale graphs, they proposed a sub-graph training strategy, which first trained the sampled sub-graph in a small batch. Based on a deep generative graph model, a novel method called DeepNC for inferring the missing parts of a network was proposed in~\cite{tran2019deepnc}.

\begin{figure*}[htb]
	\centering
	\includegraphics[width=4.5in,height=1.5in]{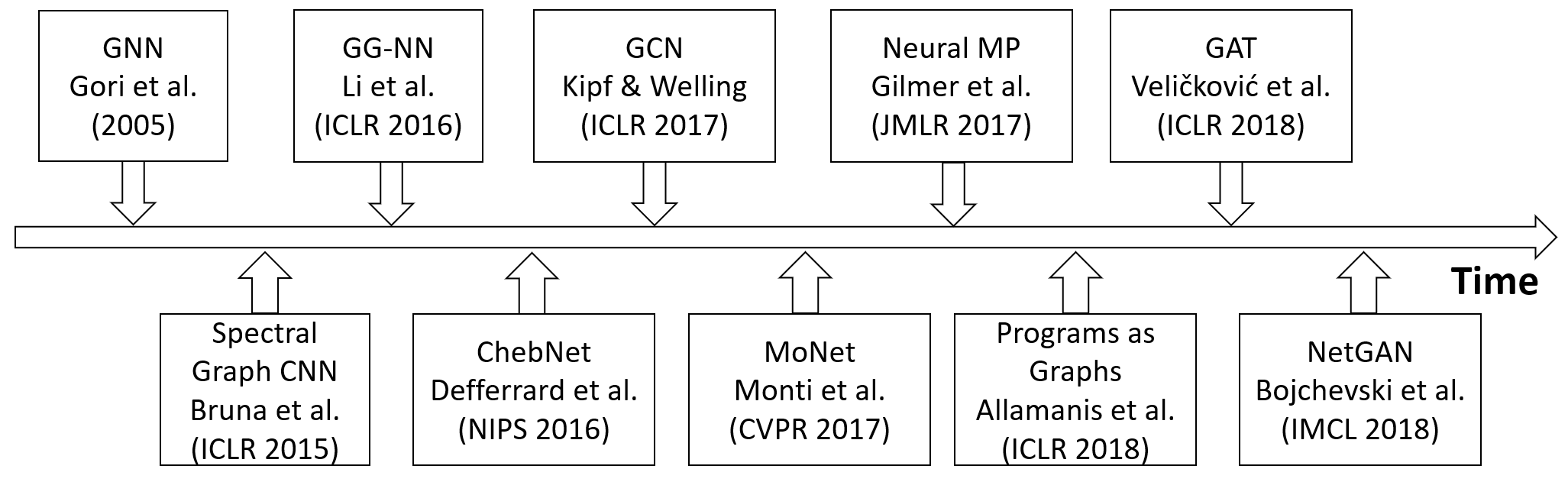}
	\caption{A brief history of algorithms of deep learning on graphs.}
	\label{fig3}
\end{figure*}

A brief history of deep learning on graphs is shown in Fig.~\ref{fig3}. GNN has attracted lots of attention since 2015, and it is widely studied and used in various fields.

\subsubsection{Graph Attention Networks}
In sequence-based tasks, attention mechanism has been regarded as a standard~\cite{vaswani2017attention}. GNNs achieve lots of benefits from the expanded model capacity of attention mechanisms. GATs are a kind of spatial-based GCNs~\cite{velivckovic2017graph}. It takes the attention mechanism into consideration when determining the weights of vertex's neighbors. Likewise, Gated Attention Networks (GAANs) also introduced the multi-head attention mechanism for updating the hidden state of some vertices~\cite{zhang2018gaan}. Unlike GATs, GAANs employ a self-attention mechanism which can compute different weights for different heads. Some other models such as graph attention model (GAM) were proposed for solving different problems~\cite{lee2018graph}. Take GAM as an example, the purpose of GAM is to handle graph classification. Therefore, GAM is set to process informative parts by visiting a sequence of significant vertices adaptively. The model of GAM contains LSTM network, and some parameters contain historical information, policies, and other information generated from exploration of the graph. Attention Walks (AWs) are another kind of learning model based on GNN and random walks~\cite{abu2018watch}. In contrast to DeepWalk, AWs use differentiable attention weights when factorizing the co-occurrence matrix~\cite{perozzi2014deepwalk}.

\subsubsection{Graph Auto-Encoders}
GAE uses GNN structure to embed network vertices into low dimensional vectors. One of the most general solutions is to employ a multi-layer perception as the encoder for inputs~\cite{hou2021sac}. Therein the decoder reconstructs neighborhood statistics of the vertex. PPMI or the first and the second nearest neighborhood can be taken into statistics~\cite{cao2016deep,wang2016structural}. Deep neural networks for graph representations (DNGR) employ PPMI. Structural deep network embedding (SDNE) employs stacked auto-encoder to maintain both the first-order and the second-order proximity. Auto-encoder~\cite{qi2014robust} is a traditional deep learning model, which can be classified as a self-supervised model~\cite{jing2020self}. Deep recursive network embedding (DRNE) reconstructs some vertices' hidden state rather than the entire graph~\cite{tu2018deep}. It has been found that if we regard GCN as an encoder, and combine GCN with GAN or LSTM with GAN, then we can design the auto-encoder for graphs. Generally speaking, DNGR and SDNE embed vertices by the given structure features, while other methods such as DRNE learn both topology structure and content features~\cite{cao2016deep,wang2016structural}. Variational graph auto-encoder~\cite{kipf2016variational} is another successful approach that employs GCN as an encoder and a link prediction layer as a decoder. Its successor, adversarially regularized variational graph auto-encoder~\cite{pan2019learning}, adds a regularization process with an adversarial training approach to learn a more robust embedding.

\subsubsection{Graph Generative Networks}
The purpose of graph generative networks is to generate graphs according to the given observed set of graphs. Many previous methods of graph generative networks have their own application domains. For example, in natural language processing, the semantic graph or the knowledge graph is generated based on the given sentences. Some general methods have been proposed recently. One kind of them considers the generation process as the formation of vertices and edges. Another kind is to employ generative adversarial training. Some GCNs based graph generative networks such as molecular generative adversarial networks (MolGAN) integrate GNN with reinforcement learning~\cite{schlichtkrull2018modeling}. Deep generative models of graphs (DGMG) achieves a hidden representation of existing graphs by utilizing spatial-based GCNs~\cite{li2018learning}. There are some knowledge graph embedding algorithms based on GAN and Zero-Shot Learning~\cite{xian2018zero}. Vyas et al.~\cite{vyas2020leveraging} proposed a Generalized Zero-Shot learning model, which can find unseen semantic in knowledge graphs.

\subsubsection{Graph Spatial-Temporal Networks}
Graph spatial-temporal networks simultaneously capture the spatial and temporal dependence of graphs. The global structure is included in the spatial-temporal graphs, and the input of each vertex varies with the change of time. For example, in traffic networks, each sensor records the traffic speed of a road continuously as a vertex, in which the edge of the traffic networks is determined by the distance between the sensor pairs~\cite{li2017diffusion}. The goal of a spatial-temporal network can be to predict future vertex values or labels, or to predict spatial-temporal graph labels. Recent studies in this direction have discussed the use of GCNs, the combination of GCNs with RNN or CNN, and recursive structures for graph structures~\cite{yu2017spatio,yan2018spatial,wu2019graph}.

\begin{table*}[htbp]
  \centering
  \caption{Summary of graph learning methods and their applications}
    \begin{tabular}{clll}
    \toprule
    \textbf{Categories} & \multicolumn{1}{c}{\textbf{Algorithms}} & \multicolumn{1}{c}{\textbf{Neural Component}} & \multicolumn{1}{c}{\textbf{Applications}} \\
    \midrule
    \multicolumn{1}{c}{\multirow{8}[16]{*}{Time Domain and\newline{} Spectral Methods}}
                       & SNLCN~\cite{bruna2013spectral} & Graph Neural Network &
                       Classification\\
\cmidrule{2-4}          & DCN~\cite{henaff2015deep} & Spectral Network  & Classification \\
\cmidrule{2-4}          & ChebNet~\cite{defferrard2016convolutional} & Convolution Network & Classification \\
\cmidrule{2-4}          & GCN~\cite{kipf2016semi} & Spectral Network  & Classification \\
\cmidrule{2-4}          & HA-GCN~\cite{zhou2017graph} & GCN   & Classification \\
\cmidrule{2-4}          & Dynamic GCN~\cite{manessi2017dynamic} & GCN, LSTM & Classification \\
\cmidrule{2-4}          & DCRNN~\cite{li2017diffusion} & Diffusion Convolution Network & Traffic Forecasting \\
\cmidrule{2-4}          & ST-GCN ~\cite{yan2018spatial} & GCN   & Action Recognition \\
    \midrule
    \multicolumn{1}{c}{\multirow{7}[14]{*}{Space Domain and\newline{} Spatial Methods}} & PATCHY-SAN~\cite{niepert2016learning} & Convolutional Network & \multicolumn{1}{p{11.94em}}{Runtime Analysis,\newline{}Feature Visualization,\newline{}Graph Classification} \\
\cmidrule{2-4}          & \multicolumn{1}{p{11.5em}}{Neural FP~\cite{duvenaud2015convolutional}} &       & Sub-graph Classification \\
\cmidrule{2-4}          & DCNN~\cite{atwood2016diffusion} & DCNN  & Classification \\
\cmidrule{2-4}          & DGCN~\cite{zhuang2018dual} & \multicolumn{1}{p{11.5em}}{Graph-Structure-Based Convolution, PPMI-Based Convolution.} & Classification \\
\cmidrule{2-4}          & SSE~\cite{dai2018learning} &       & Vertex Classification \\
\cmidrule{2-4}          & LGCN~\cite{gao2018large} & Convolutional Neural Network & Vertex Classification \\
\cmidrule{2-4}          & STGCN~\cite{yu2017spatio} & Gated Sequential Convolution & Traffic Forecasting \\
    \midrule
    \multicolumn{1}{l}{\multirow{12}[24]{*}{Deep Learning\newline{}  Model Based Methods}} & GATs~\cite{velivckovic2017graph} & \multicolumn{1}{l}{\multirow{3}[6]{*}{Attention Neural Network}} & Classification \\
\cmidrule{2-2}\cmidrule{4-4}          & GAAN~\cite{zhang2018gaan} &       & Vertex Classification \\
\cmidrule{2-2}\cmidrule{4-4}          & GAM~\cite{lee2018graph} &       & Graph Classification \\
\cmidrule{2-4}          & Aws~\cite{abu2018watch} & \multicolumn{1}{l}{\multirow{4}[8]{*}{Auto-encoder Neural Network}} & \multicolumn{1}{p{11.94em}}{Link Prediction,  \newline{}Sensitivity Analysis,\newline{}Vertex Classification} \\
\cmidrule{2-2}\cmidrule{4-4}          & SDNE~\cite{wang2016structural} &       & \multicolumn{1}{p{11.94em}}{Classification,\newline{} Link Prediction, \newline{}Visualization} \\
\cmidrule{2-2}\cmidrule{4-4}          & DNGR~\cite{cao2016deep}  &       & Clustering, Visualization \\
\cmidrule{2-2}\cmidrule{4-4}          & DRNE~\cite{tu2018deep}&       & \multicolumn{1}{p{11.94em}}{Regular Equivalence Prediction,\newline{}Structural Role Classification,\newline{}Network Visualization} \\
\cmidrule{2-4}          & MolGAN~\cite{schlichtkrull2018modeling}  & \multicolumn{1}{l}{\multirow{2}[4]{*}{Generative Neural Network}} & Generative Model \\
\cmidrule{2-2}\cmidrule{4-4}          & DGMG~\cite{li2018learning} &       &  Molecule Generation \\
\cmidrule{2-4}          & DCRNN~\cite{li2017diffusion} & Diffusion Convolution Network & Traffic Forecasting \\
\cmidrule{2-4}          & STGCN~\cite{yu2017spatio} & Gated Sequential Convolution &  \\
\cmidrule{2-4}          & ST-GCN~\cite{yan2018spatial} & GCNs  & Action Recognition \\
    \bottomrule
    \end{tabular}%
  \label{tab:addlabel}%
\end{table*}%

\subsubsection{Discussion}
In this context, the task of graph learning can be seen as optimizing the objective function by using gradient descent algorithms. Therefore the performance of deep learning based NRL models is influenced by gradient descent algorithms. They may encounter challenges like local optimal solutions and the vanishing gradient problem.

\section{Applications}\label{sec:appication}

Many problems can be solved by graph learning methods, including supervised, semi-supervised, unsupervised, and reinforcement learning. Some researchers classify the applications of graph learning into three categories, i.e., structural scenarios, non-structural scenarios, and other application scenarios~\cite{zhang2018deep}. Structural scenarios refer to the situation where data are performed in explicit relational structures, such as physical systems, molecular structures, and knowledge graphs. Non-structural scenarios refer to the situation where data are with unclear relational structures, such as images and texts. Other application scenarios include, e.g., integrating models and combinatorial optimization problems. Table II lists the neural components and applications of various graph learning methods.

\subsection{Datasets and Open-source Libraries}
There are several datasets and benchmarks used to evaluate the performance of graph learning approaches for various tasks such as link prediction, node classification, and graph visualization. For instance, datasets like Cora\footnote{https://relational.fit.cvut.cz/dataset/CORA} (citation network), Pubmed\footnote{https://catalog.data.gov/dataset/pubmed} (citation network), BlogCatalog\footnote{http://networkrepository.com/soc-BlogCatalog.php} (social network), Wikipedia\footnote{https://en.wikipedia.org/wiki/Wikipedia:Database\_download} (language network) and PPI\footnote{https://openwetware.org/wiki/Protein-protein\_interaction\_databases} (biological network) include nodes, edges, labels or attributes of nodes. Some research institutions developed graph learning libraries, which include common and classical graph learning algorithms. For example, OpenKE\footnote{https://github.com/thunlp/OpenKE} is a Python library for knowledge graph embedding based on PyTorch. The open-source framework has the implementations of RESCAL, HolE, DistMult, ComplEx, etc. CogDL\footnote{https://github.com/THUDM/cogdl/} is a graph representation learning framework, which can be used for node classification, link prediction, graph classification, etc.

\subsection{Text}
Many data are in textual form coming from various resources like web pages, emails, documents (technical and corporate), books, digital libraries and customer complains, letters, patents, etc. Textual data are not well structured for obtaining any meaningful information as text often contains rich context information. There exist abundant applications around text, including text classification, sequence labeling, sentiment classification, etc. Text classification is one of the most classical problems in natural language processing. Popular algorithms proposed to handle this problem include GCNs~\cite{hamilton2017inductive, kipf2016semi}, GATs~\cite{velivckovic2017graph}, Text GCNs~\cite{yao2019graph}, and Sentence LSTM~\cite{zhang2018sentence}. Sentence LSTM has also been applied to sequence labeling, text generation, multi-hop reading comprehension, etc~\cite{zhang2018sentence}. Syntactic GCN was proposed to solve semantic role labeling and neural machine translation~\cite{marcheggiani2017encoding}. Gated Graph Neural Networks (GGNNs) can also be used to address neural machine translation and text generation~\cite{beck2018graph}. For relational extraction, Tree LSTM, graph LSTM, and GCN are better solutions~\cite{peng2018large}.

\subsection{Images}
Graph learning applications pertaining to images include social relationship understanding, image classification, visual question answering, object detection, region classification, and semantic segmentation, etc. For social relationship understanding, for instance, graph reasoning model (GRM) is widely used~\cite{wang2018deep}. Since social relationships such as friendships are the basis of social networks in real world, automatically interpreting these relationships is important for understanding human behaviors. GRM introduces GGNNs to learn a propagation mechanism. Image classification is a classical problem, in which GNNs have demonstrated promising performance. Visual question answering (VQA) is a learning task that involves both computer vision and natural language processing. A VQA system takes the form of a certain pictures and its open natural language question as input, in order to generate a natural language answer as output. Generally speaking, VQA is question-and-answer for a given picture. GGNNs have been exploited to help with VQA~\cite{lee2018multi}.

\subsection{Science}
Graph learning has been widely adopted in science. Modeling real-world physical systems is one of the most fundamental perspectives in understanding human intelligence. Representing objects as vertices and relations as edges between them is a simple but effective way to perform physics. Battaglia et al.~\cite{battaglia2016interaction} proposed interaction networks (IN) to predict and infer abundant physical systems, in which IN takes objects and relationships as input. Based on IN, the interactions can be reasoned and the effects can be applied. Therefore, physical dynamics can be predicted. Visual interaction networks (VIN) can make predictions from pixels by firstly learning a state code from two continuous input frames per object~\cite{watters2017visual}.

Other graph networks based models have been developed to address chemistry and biology problems. Calculating molecular fingerprints, i.e., using feature vectors to represent molecular, is a central step. Researchers~\cite{butler2018machine} proposed neural graph fingerprints using GCNs to calculate substructure feature vectors. Some studies focused on protein interface prediction. This is a challenging issue with significant applications in biology. Besides, GNNs can be used in biomedical engineering as well. Based on protein-protein interaction networks, Rhee et al.~\cite{rhee2017hybrid} used graph convolution and protein relation networks to classify breast cancer subtypes.

\subsection{Knowledge Graphs}
Various heterogeneous objects and relationships are regarded as the basis for a knowledge graph \cite{ji2020survey}. GNNs can be applied in knowledge base completion (KBC) for solving the out-of-knowledge-base (OOKB) entity problem~\cite{hamaguchi2018knowledge}. The OOKB entities are connected to  existing entities. Therefore, the embedding of OOKB entities can be aggregated from existing entities. Such kind of algorithms achieve reasonable performance in both settings of KBC and OOKB. Likewise, GCNs can also be used to solve the problem of cross-lingual knowledge graph alignment. The main idea of the model is to embed entities from different languages into an integrated embedding space. Then the model aligns these entities according to their embedding similarities.

Generally speaking, knowledge graph embedding can be categorized into two types: translational distance models and semantic matching models. Translational distance models aim to learn the low dimensional vector of entities in a knowledge graph by employing distance-based scoring functions. These methods calculate the plausibility as the distance between two entities after a translation measured by the relationships between them. Among current translational distance models, TransE~\cite{bordes2013translating} is the most influential one. TransE can model the relationship of entities by interpreting them as translations operating on the low dimensional embedding. Inspired by TranE, TranH~\cite{wang2014knowledge} was proposed to overcome the disadvantages of TransE in dealing with 1-to-N, N-to-1, and N-to-N relations by introducing relation-specific hyperplanes. Instead of hyperplanes, TransR~\cite{lin2015learning} introduces relation-specific spaces to solve the flows in TransE. Meanwhile, various extensions of TransE have been proposed to enhance knowledge graph embeddings, such as TransD~\cite{ji2015knowledge} and TransF~\cite{feng2016knowledge}. On the basis of TransE, DeepPath \cite{huang2017heterogeneous} incorporates reinforcement learning methods for learning relational paths in knowledge graphs. By designing a complex reward function involving accuracy, efficiency and path diversity, the path finding process is better controlled and more flexible.

Semantic matching models utilize the similarity-based scoring functions. They measure the plausibility among entities by matching latent semantics of entities and relations in low dimensional vector space. Typical models of this type include RESCAL~\cite{jenatton2012latent}, DistMult~\cite{yang2014embedding}, ANALOGY~\cite{liu2017analogical}, etc.

\subsection{Combinatorial Optimization}
 Classical problems such as traveling salesman problem (TSP) and minimum spanning tree (MST) have been solved by using different heuristic solutions. Recently, deep neural networks have been applied to these problems. Some solutions make further use of GNNs thanks to their structures. Bello et al.~\cite{bello2016neural} first proposed such kind of methods to solve TSP. Their method mainly contains two steps, i.e., a parameterized reward pointer network and a strategy gradient module for training. Khalil et al.~\cite{khalil2017learning} improved this work with GNN and achieved better performance by two main procedures. First, they used structure2vec to achieve vertex embedding and then input them into Q-learning module for decision-making. This work also proves the embedding ability of GNN. Nowak et al. ~\cite{nowak2018revised} focused on the secondary assignment problem, i.e., measuring the similarity of two graphs. The GNN model learns each graph's vertex embedding and uses the attention mechanism to match the two graphs. Other studies use GNNs directly as the classifiers, which can perform the intensive prediction on graphs with two sides. The rest of the model facilitates diverse choices and effective training.

\section{Open Issues}\label{sec:issues}
In this section, we briefly summarize several future research directions and open issues for graph learning.

\textbf{Dynamic Graph Learning}: For the purpose of graph learning, most existing algorithms are suitable for static networks without specific constraints. However, dynamic networks such as traffic networks vary over time. Therefore, they are hard to deal with. Dynamic graph learning algorithms have rarely been studied in the literature. It is of significant importance that dynamic graph learning algorithms are designed to maintain good performance, especially in the case of dynamic graphs.

\textbf{Generative Graph Learning}: Inspired by the generative adversarial networks, generative graph learning algorithms can unify the generative and discriminative models by playing a game-theoretical min-max game. This generative graph learning method can be used for link prediction, network evolution, and recommendation by boosting the performance of generative and discriminative models alternately and iteratively.

\textbf{Fair Graph Learning}: Most graph learning algorithms rely on deep neural networks, and the resulting vectors may have captured undesired sensitive information. The bias existing in the network is reinforced, and hence it is of significant importance to integrate the fair metrics into the graph learning algorithms to address the inherent bias issue.

\textbf{Interpretability of Graph Learning}: The models of graph learning are generally complex by incorporating both graph structure and feature information. The interpretability of graph learning (based) algorithms remains unsolved since the structures of graph learning algorithms are still a black box. For example, drug discovery can be achieved by graph learning algorithms. However, it is unknown how this drug is discovered as well as the reason behind this discovery. The interpretability behind graph learning needs to be further studied.

\section{Conclusion}\label{sec:con}
This survey gives a general description of graph learning, and provides a comprehensive review of the state-of-the-art graph learning methods. We examined existing graph learning methods under four categories: graph signal processing based methods, matrix factorization based methods, random walk based methods, and deep learning based methods. The applications of graph learning methods mainly under these four categories in areas such as text, images, science, knowledge graphs, and combinatorial optimization are outlined. We also discuss some future research directions in the field of graph learning. Graph learning is currently a hot area which is growing at an unprecedented speed. We do hope that this survey will help researchers and practitioners with their research and development in graph learning and related areas.

\section*{Acknowledgments}
The authors would like to thank Prof. Hussein Abbass at University of New South Wales, Yuchen Sun, Jiaying Liu, Hao Ren at Dalian University of Technology, and anonymous reviewers for their valuable comments and suggestions.

\ifCLASSOPTIONcaptionsoff
  \newpage
\fi
\bibliographystyle{IEEEtran}
\bibliography{GraphLearningSurvey}

\begin{thebibliography}{100}
\providecommand{\url}[1]{#1}
\csname url@samestyle\endcsname
\providecommand{\newblock}{\relax}
\providecommand{\bibinfo}[2]{#2}
\providecommand{\BIBentrySTDinterwordspacing}{\spaceskip=0pt\relax}
\providecommand{\BIBentryALTinterwordstretchfactor}{4}
\providecommand{\BIBentryALTinterwordspacing}{\spaceskip=\fontdimen2\font plus
\BIBentryALTinterwordstretchfactor\fontdimen3\font minus
  \fontdimen4\font\relax}
\providecommand{\BIBforeignlanguage}[2]{{%
\expandafter\ifx\csname l@#1\endcsname\relax
\typeout{** WARNING: IEEEtran.bst: No hyphenation pattern has been}%
\typeout{** loaded for the language `#1'. Using the pattern for}%
\typeout{** the default language instead.}%
\else
\language=\csname l@#1\endcsname
\fi
#2}}
\providecommand{\BIBdecl}{\relax}
\BIBdecl

\bibitem{fortunato2018science}
S.~Fortunato, C.~T. Bergstrom, K.~B{\"o}rner, J.~A. Evans, D.~Helbing,
  S.~Milojevi{\'c}, A.~M. Petersen, F.~Radicchi, R.~Sinatra, B.~Uzzi
  \emph{et~al.}, ``Science of science,'' \emph{Science}, vol. 359, no. 6379,
  2018, eaao0185.

\bibitem{liu2020wos}
J.~Liu, J.~Ren, W.~Zheng, L.~Chi, I.~Lee, and F.~Xia, ``Web of scholars: A
  scholar knowledge graph,'' in \emph{the 43rd International ACM SIGIR
  Conference on Research and Development in Information Retrieval (SIGIR)},
  2020, pp. 2153--2156.

\bibitem{liu2019shifu2}
J.~Liu, F.~Xia, L.~Wang, B.~Xu, X.~Kong, H.~Tong, and I.~King, ``Shifu2: A
  network representation learning based model for advisor-advisee relationship
  mining,'' \emph{IEEE Transactions on Knowledge and Data Engineering},
  vol.~33, no.~4, pp. 1763--1777, 2021.

\bibitem{zhang2018network}
D.~Zhang, J.~Yin, X.~Zhu, and C.~Zhang, ``Network representation learning: A
  survey,'' \emph{IEEE Transactions on Big Data}, vol.~6, no.~1, pp. 3--28,
  2020.

\bibitem{sun2020bigdata}
K.~Sun, J.~Liu, S.~Yu, B.~Xu, and F.~Xia, ``Graph force learning,'' in
  \emph{IEEE International Conference on Big Data (BigData)}, 2020, pp.
  2987--2994.

\bibitem{xia2020TITSstation}
F.~Xia, J.~Wang, X.~Kong, D.~Zhang, and Z.~Wang, ``Ranking station importance
  with human mobility patterns using subway network datasets,'' \emph{IEEE
  Transactions on Intelligent Transportation Systems}, vol.~21, no.~7, pp.
  2840--2852, 2020.

\bibitem{grover2016node2vec}
A.~Grover and J.~Leskovec, ``node2vec: Scalable feature learning for
  networks,'' in \emph{Proceedings of the 22nd ACM SIGKDD International
  Conference on Knowledge Discovery and Data Mining}.\hskip 1em plus 0.5em
  minus 0.4em\relax ACM, 2016, pp. 855--864.

\bibitem{nrl2020access}
K.~Sun, L.~Wang, B.~Xu, W.~Zhao, S.~W. Teng, and F.~Xia, ``Network
  representation learning: From traditional feature learning to deep
  learning,'' \emph{IEEE Access}, vol.~8, no.~1, pp. 205\,600--205\,617, 2020.

\bibitem{yu2020cikm}
S.~Yu, F.~Xia, J.~Xu, Z.~Chen, and I.~Lee, ``Offer: A motif dimensional
  framework for network representation learning,'' in \emph{The 29th ACM
  International Conference on Information and Knowledge Management (CIKM)},
  2020, pp. 3349--3352.

\bibitem{bengio2013representation}
Y.~Bengio, A.~Courville, and P.~Vincent, ``Representation learning: A review
  and new perspectives,'' \emph{IEEE Transactions on Pattern Analysis and
  Machine Intelligence}, vol.~35, no.~8, pp. 1798--1828, 2013.

\bibitem{guo2020aaai}
T.~Guo, F.~Xia, S.~Zhen, X.~Bai, D.~Zhang, Z.~Liu, and J.~Tang, ``Graduate
  employment prediction with bias,'' in \emph{Thirty-Fourth AAAI Conference on
  Artificial Intelligence (AAAI)}, 2020, pp. 670--677.

\bibitem{xia2014exploiting}
F.~Xia, A.~M. Ahmed, L.~T. Yang, J.~Ma, and J.~J. Rodrigues, ``Exploiting
  social relationship to enable efficient replica allocation in ad-hoc social
  networks,'' \emph{IEEE Transactions on Parallel and Distributed Systems},
  vol.~25, no.~12, pp. 3167--3176, 2014.

\bibitem{ddcss2020bdr}
J.~Zhang, W.~Wang, F.~Xia, Y.-R. Lin, and H.~Tong, ``Data-driven computational
  social science: A survey,'' \emph{Big Data Research}, vol.~21, p. 100145,
  2020.

\bibitem{xia2014community}
F.~Xia, A.~M. Ahmed, L.~T. Yang, and Z.~Luo, ``Community-based event
  dissemination with optimal load balancing,'' \emph{IEEE Transactions on
  Computers}, vol.~64, no.~7, pp. 1857--1869, 2014.

\bibitem{motif2019access}
F.~Xia, H.~Wei, S.~Yu, D.~Zhang, and B.~Xu, ``A survey of measures for network
  motifs,'' \emph{IEEE Access}, vol.~7, no.~1, pp. 106\,576--106\,587, 2019.

\bibitem{leskovec2006sampling}
J.~Leskovec and C.~Faloutsos, ``Sampling from large graphs,'' in
  \emph{Proceedings of the 12th ACM SIGKDD International Conference on
  Knowledge Discovery and Data Mining}.\hskip 1em plus 0.5em minus 0.4em\relax
  ACM, 2006, pp. 631--636.

\bibitem{wang2015local}
X.~Wang, P.~Liu, and Y.~Gu, ``Local-set-based graph signal reconstruction,''
  \emph{IEEE Transactions on Signal Processing}, vol.~63, no.~9, pp.
  2432--2444, 2015.

\bibitem{zhang2018deep}
Z.~Zhang, P.~Cui, and W.~Zhu, ``Deep learning on graphs: A survey,'' \emph{IEEE
  Transactions on Knowledge and Data Engineering}, 2020.

\bibitem{xu2020jcdl}
J.~Xu, S.~Yu, K.~Sun, J.~Ren, I.~Lee, S.~Pan, and F.~Xia, ``Multivariate
  relations aggregation learning in social networks,'' in \emph{ACM/IEEE Joint
  Conference on Digital Libraries (JCDL)}, 2020, pp. 77--86.

\bibitem{bedru2020big}
H.~D. Bedru, S.~Yu, X.~Xiao, D.~Zhang, L.~Wan, H.~Guo, and F.~Xia, ``Big
  networks: A survey,'' \emph{Computer Science Review}, vol.~37, p. 100247,
  2020.

\bibitem{goyal2018graph}
P.~Goyal and E.~Ferrara, ``Graph embedding techniques, applications, and
  performance: A survey,'' \emph{Knowledge-Based Systems}, vol. 151, pp.
  78--94, 2018.

\bibitem{cai2018comprehensive}
H.~Cai, V.~W. Zheng, and K.~C.-C. Chang, ``A comprehensive survey of graph
  embedding: Problems, techniques, and applications,'' \emph{IEEE Transactions
  on Knowledge and Data Engineering}, vol.~30, no.~9, pp. 1616--1637, 2018.

\bibitem{nickel2016review}
M.~Nickel, K.~Murphy, V.~Tresp, and E.~Gabrilovich, ``A review of relational
  machine learning for knowledge graphs,'' \emph{Proceedings of the IEEE}, vol.
  104, no.~1, pp. 11--33, 2016.

\bibitem{akoglu2015graph}
L.~Akoglu, H.~Tong, and D.~Koutra, ``Graph based anomaly detection and
  description: a survey,'' \emph{Data Mining and Knowledge Discovery}, vol.~29,
  no.~3, pp. 626--688, 2015.

\bibitem{wu2019comprehensive}
Z.~Wu, S.~Pan, F.~Chen, G.~Long, C.~Zhang, and P.~S. Yu, ``A comprehensive
  survey on graph neural networks,'' \emph{IEEE Transactions on Neural Networks
  and Learning Systems}, vol.~32, no.~1, pp. 4--24, 2021.

\bibitem{ortega2018graph}
A.~Ortega, P.~Frossard, J.~Kova{\v{c}}evi{\'c}, J.~M. Moura, and
  P.~Vandergheynst, ``Graph signal processing: Overview, challenges, and
  applications,'' \emph{Proceedings of the IEEE}, vol. 106, no.~5, pp.
  808--828, 2018.

\bibitem{huang2018graph}
W.~Huang, T.~A. Bolton, J.~D. Medaglia, D.~S. Bassett, A.~Ribeiro, and D.~Van
  De~Ville, ``A graph signal processing perspective on functional brain
  imaging,'' \emph{Proceedings of the IEEE}, vol. 106, no.~5, pp. 868--885,
  2018.

\bibitem{xia2017big}
F.~Xia, W.~Wang, T.~M. Bekele, and H.~Liu, ``Big scholarly data: A survey,''
  \emph{IEEE Transactions on Big Data}, vol.~3, no.~1, pp. 18--35, 2017.

\bibitem{lecun2015deep}
Y.~LeCun, Y.~Bengio, and G.~Hinton, ``Deep learning,'' \emph{Nature}, vol. 521,
  no. 7553, p. 436, 2015.

\bibitem{liu2018artificial}
J.~Liu, X.~Kong, F.~Xia, X.~Bai, L.~Wang, Q.~Qing, and I.~Lee, ``Artificial
  intelligence in the 21st century,'' \emph{IEEE Access}, vol.~6, pp.
  34\,403--34\,421, 2018.

\bibitem{sandryhaila2013discrete}
A.~Sandryhaila and J.~M. Moura, ``Discrete signal processing on graphs,''
  \emph{IEEE Transactions on Signal Processing}, vol.~61, no.~7, pp.
  1644--1656, 2013.

\bibitem{shuman2012emerging}
D.~Shuman, S.~Narang, P.~Frossard, A.~Ortega, and P.~Vandergheynst, ``The
  emerging field of signal processing on graphs: Extending high-dimensional
  data analysis to networks and other irregular domains,'' \emph{IEEE Signal
  Processing Magazine}, vol.~3, no.~30, pp. 83--98, 2013.

\bibitem{puschel2006algebraic}
M.~Puschel and J.~M. Moura, ``Algebraic signal processing theory: Foundation
  and 1-d time,'' \emph{IEEE Transactions on Signal Processing}, vol.~56,
  no.~8, pp. 3572--3585, 2008.

\bibitem{sandryhaila2013discrete1}
A.~Sandryhaila and J.~M. Moura, ``Discrete signal processing on graphs: Graph
  filters,'' in \emph{2013 IEEE International Conference on Acoustics, Speech
  and Signal Processing}.\hskip 1em plus 0.5em minus 0.4em\relax IEEE, 2013,
  pp. 6163--6166.

\bibitem{chen2015discrete}
S.~Chen, R.~Varma, A.~Sandryhaila, and J.~Kova{\v{c}}evi{\'c}, ``Discrete
  signal processing on graphs: Sampling theory,'' \emph{IEEE Transactions on
  Signal Processing}, vol.~63, no.~24, pp. 6510--6523, 2015.

\bibitem{von2007tutorial}
U.~Von~Luxburg, ``A tutorial on spectral clustering,'' \emph{Statistics and
  Computing}, vol.~17, no.~4, pp. 395--416, 2007.

\bibitem{zhu2012graph}
X.~Zhu and M.~Rabbat, ``Graph spectral compressed sensing for sensor
  networks,'' in \emph{2012 IEEE International Conference on Acoustics, Speech
  and Signal Processing (ICASSP)}.\hskip 1em plus 0.5em minus 0.4em\relax IEEE,
  2012, pp. 2865--2868.

\bibitem{gavili2017shift}
A.~Gavili and X.-P. Zhang, ``On the shift operator, graph frequency, and
  optimal filtering in graph signal processing,'' \emph{IEEE Transactions on
  Signal Processing}, vol.~65, no.~23, pp. 6303--6318, 2017.

\bibitem{pasdeloup2015graph}
B.~Pasdeloup, M.~Rabbat, V.~Gripon, D.~Pastor, and G.~Mercier, ``Graph
  reconstruction from the observation of diffused signals,'' in \emph{2015 53rd
  Annual Allerton Conference on Communication, Control, and Computing
  (Allerton)}.\hskip 1em plus 0.5em minus 0.4em\relax IEEE, 2015, pp.
  1386--1390.

\bibitem{anis2014towards}
A.~Anis, A.~Gadde, and A.~Ortega, ``Towards a sampling theorem for signals on
  arbitrary graphs,'' in \emph{2014 IEEE International Conference on Acoustics,
  Speech and Signal Processing (ICASSP)}.\hskip 1em plus 0.5em minus
  0.4em\relax IEEE, 2014, pp. 3864--3868.

\bibitem{puy2018random}
G.~Puy, N.~Tremblay, R.~Gribonval, and P.~Vandergheynst, ``Random sampling of
  bandlimited signals on graphs,'' \emph{Applied and Computational Harmonic
  Analysis}, vol.~44, no.~2, pp. 446--475, 2018.

\bibitem{shomorony2014sampling}
H.~Shomorony and A.~S. Avestimehr, ``Sampling large data on graphs,'' in
  \emph{2014 IEEE Global Conference on Signal and Information Processing
  (GlobalSIP)}.\hskip 1em plus 0.5em minus 0.4em\relax IEEE, 2014, pp.
  933--936.

\bibitem{chamon2018greedy}
L.~F. Chamon and A.~Ribeiro, ``Greedy sampling of graph signals,'' \emph{IEEE
  Transactions on Signal Processing}, vol.~66, no.~1, pp. 34--47, 2018.

\bibitem{marques2016sampling}
A.~G. Marques, S.~Segarra, G.~Leus, and A.~Ribeiro, ``Sampling of graph signals
  with successive local aggregations.'' \emph{IEEE Transactions Signal
  Processing}, vol.~64, no.~7, pp. 1832--1843, 2016.

\bibitem{narang2013signal}
S.~K. Narang, A.~Gadde, and A.~Ortega, ``Signal processing techniques for
  interpolation in graph structured data,'' in \emph{2013 IEEE International
  Conference on Acoustics, Speech and Signal Processing}.\hskip 1em plus 0.5em
  minus 0.4em\relax IEEE, 2013, pp. 5445--5449.

\bibitem{gadde2015probabilistic}
A.~Gadde and A.~Ortega, ``A probabilistic interpretation of sampling theory of
  graph signals,'' in \emph{2015 IEEE International Conference on Acoustics,
  Speech and Signal Processing (ICASSP)}.\hskip 1em plus 0.5em minus
  0.4em\relax IEEE, 2015, pp. 3257--3261.

\bibitem{wang2015distributed}
X.~Wang, M.~Wang, and Y.~Gu, ``A distributed tracking algorithm for
  reconstruction of graph signals,'' \emph{IEEE Journal of Selected Topics in
  Signal Processing}, vol.~9, no.~4, pp. 728--740, 2015.

\bibitem{di2016adaptive}
P.~Di~Lorenzo, S.~Barbarossa, P.~Banelli, and S.~Sardellitti, ``Adaptive least
  mean squares estimation of graph signals,'' \emph{IEEE Transactions on Signal
  and Information Processing over Networks}, vol.~2, no.~4, pp. 555--568, 2016.

\bibitem{romero2017kernel}
D.~Romero, M.~Ma, and G.~B. Giannakis, ``Kernel-based reconstruction of graph
  signals.'' \emph{IEEE Transactions Signal Processing}, vol.~65, no.~3, pp.
  764--778, 2017.

\bibitem{nagahara2015discrete}
M.~Nagahara, ``Discrete signal reconstruction by sum of absolute values,''
  \emph{IEEE Signal Processing Letters}, vol.~22, no.~10, pp. 1575--1579, 2015.

\bibitem{chen2015signal}
S.~Chen, R.~Varma, A.~Singh, and J.~Kova{\v{c}}evi{\'c}, ``Signal
  representations on graphs: Tools and applications,'' \emph{arXiv preprint
  arXiv:1512.05406}, 2015.

\bibitem{segarra2016reconstruction}
S.~Segarra, A.~G. Marques, G.~Leus, and A.~Ribeiro, ``Reconstruction of graph
  signals through percolation from seeding nodes,'' \emph{IEEE Transactions on
  Signal Processing}, vol.~64, no.~16, pp. 4363--4378, 2016.

\bibitem{tn2020sigweb}
F.~Xia, J.~Liu, J.~Ren, W.~Wang, and X.~Kong, ``Turing number: How far are you
  to a. m. turing award?'' \emph{ACM SIGWEB Newsletter}, vol. Autumn, 2020,
  article No.: 5.

\bibitem{egilmez2017graph}
H.~E. Egilmez, E.~Pavez, and A.~Ortega, ``Graph learning from data under
  laplacian and structural constraints,'' \emph{IEEE Journal of Selected Topics
  in Signal Processing}, vol.~11, no.~6, pp. 825--841, 2017.

\bibitem{dong2016learning}
X.~Dong, D.~Thanou, P.~Frossard, and P.~Vandergheynst, ``Learning laplacian
  matrix in smooth graph signal representations,'' \emph{IEEE Transactions on
  Signal Processing}, vol.~64, no.~23, pp. 6160--6173, 2016.

\bibitem{kalofolias2016learn}
V.~Kalofolias, ``How to learn a graph from smooth signals,'' in
  \emph{Artificial Intelligence and Statistics}, 2016, pp. 920--929.

\bibitem{pavez2016generalized}
E.~Pavez and A.~Ortega, ``Generalized laplacian precision matrix estimation for
  graph signal processing,'' in \emph{2016 IEEE International Conference on
  Acoustics, Speech and Signal Processing (ICASSP)}.\hskip 1em plus 0.5em minus
  0.4em\relax IEEE, 2016, pp. 6350--6354.

\bibitem{pavez2018learning}
E.~Pavez, H.~E. Egilmez, and A.~Ortega, ``Learning graphs with monotone
  topology properties and multiple connected components,'' \emph{IEEE
  Transactions on Signal Processing}, vol.~66, no.~9, pp. 2399--2413, 2018.

\bibitem{pasdeloup2018characterization}
B.~Pasdeloup, V.~Gripon, G.~Mercier, D.~Pastor, and M.~G. Rabbat,
  ``Characterization and inference of graph diffusion processes from
  observations of stationary signals,'' \emph{IEEE Transactions on Signal and
  Information Processing over Networks}, vol.~4, no.~3, pp. 481--496, 2018.

\bibitem{segarra2017network}
S.~Segarra, A.~G. Marques, G.~Mateos, and A.~Ribeiro, ``Network topology
  inference from spectral templates,'' \emph{IEEE Transactions on Signal and
  Information Processing over Networks}, vol.~3, no.~3, pp. 467--483, 2017.

\bibitem{thanou2017learning}
D.~Thanou, X.~Dong, D.~Kressner, and P.~Frossard, ``Learning heat diffusion
  graphs,'' \emph{IEEE Transactions on Signal and Information Processing over
  Networks}, vol.~3, no.~3, pp. 484--499, 2017.

\bibitem{mei2016signal}
J.~Mei and J.~M. Moura, ``Signal processing on graphs: Causal modeling
  ofunstructured data,'' \emph{IEEE Transactions on Signal Processing},
  vol.~65, no.~8, pp. 2077--2092, 2016.

\bibitem{segarra2017blind}
S.~Segarra, G.~Mateos, A.~G. Marques, and A.~Ribeiro, ``Blind identification of
  graph filters,'' \emph{IEEE Transactions on Signal Processing}, vol.~65,
  no.~5, pp. 1146--1159, 2017.

\bibitem{xia2013recAccess}
F.~Xia, N.~Y. Asabere, A.~M. Ahmed, J.~Li, and X.~Kong, ``Mobile multimedia
  recommendation in smart communities: A survey,'' \emph{IEEE Access}, vol.~1,
  no.~1, pp. 606--624, 2013.

\bibitem{huang2018rating}
W.~Huang, A.~G. Marques, and A.~R. Ribeiro, ``Rating prediction via graph
  signal processing,'' \emph{IEEE Transactions on Signal Processing}, vol.~66,
  no.~19, pp. 5066--5081, 2018.

\bibitem{xia2016scientific}
F.~Xia, H.~Liu, I.~Lee, and L.~Cao, ``Scientific article recommendation:
  Exploiting common author relations and historical preferences,'' \emph{IEEE
  Transactions on Big Data}, vol.~2, no.~2, pp. 101--112, 2016.

\bibitem{he2004locality}
X.~He and P.~Niyogi, ``Locality preserving projections,'' in \emph{Advances in
  Neural Information Processing Systems}, 2004, pp. 153--160.

\bibitem{chen2014unified}
M.~Chen, I.~W. Tsang, M.~Tan, and T.~J. Cham, ``A unified feature selection
  framework for graph embedding on high dimensional data,'' \emph{IEEE
  Transactions on Knowledge and Data Engineering}, vol.~27, no.~6, pp.
  1465--1477, 2014.

\bibitem{yan2007graph}
S.~Yan, D.~Xu, B.~Zhang, H.-J. Zhang, Q.~Yang, and S.~Lin, ``Graph embedding
  and extensions: A general framework for dimensionality reduction,''
  \emph{IEEE Transactions on Pattern Analysis \& Machine Intelligence}, no.~1,
  pp. 40--51, 2007.

\bibitem{borg2003modern}
I.~Borg and P.~Groenen, ``Modern multidimensional scaling: Theory and
  applications,'' \emph{Journal of Educational Measurement}, vol.~40, no.~3,
  pp. 277--280, 2003.

\bibitem{balasubramanian2002isomap}
M.~Balasubramanian and E.~L. Schwartz, ``The isomap algorithm and topological
  stability,'' \emph{Science}, vol. 295, no. 5552, pp. 7--7, 2002.

\bibitem{anderson1985eigenvalues}
W.~N. Anderson~Jr and T.~D. Morley, ``Eigenvalues of the laplacian of a
  graph,'' \emph{Linear and Multilinear Algebra}, vol.~18, no.~2, pp. 141--145,
  1985.

\bibitem{roweis2000nonlinear}
S.~T. Roweis and L.~K. Saul, ``Nonlinear dimensionality reduction by locally
  linear embedding,'' \emph{Science}, vol. 290, no. 5500, pp. 2323--2326, 2000.

\bibitem{jiang2016dimensionality}
R.~Jiang, W.~Fu, L.~Wen, S.~Hao, and R.~Hong, ``Dimensionality reduction on
  anchorgraph with an efficient locality preserving projection,''
  \emph{Neurocomputing}, vol. 187, pp. 109--118, 2016.

\bibitem{wan2019your}
L.~Wan, Y.~Yuan, F.~Xia, and H.~Liu, ``To your surprise: Identifying
  serendipitous collaborators,'' \emph{IEEE Transactions on Big Data}, 2019.

\bibitem{yang2010local}
Y.~Yang, F.~Nie, S.~Xiang, Y.~Zhuang, and W.~Wang, ``Local and global
  regressive mapping for manifold learning with out-of-sample extrapolation,''
  in \emph{Twenty-Fourth AAAI Conference on Artificial Intelligence}, 2010, pp.
  649--654.

\bibitem{xiang2008nonlinear}
S.~Xiang, F.~Nie, C.~Zhang, and C.~Zhang, ``Nonlinear dimensionality reduction
  with local spline embedding,'' \emph{IEEE Transactions on Knowledge and Data
  Engineering}, vol.~21, no.~9, pp. 1285--1298, 2008.

\bibitem{cai2007spectral}
D.~Cai, X.~He, and J.~Han, ``Spectral regression: A unified subspace learning
  framework for content-based image retrieval,'' in \emph{Proceedings of the
  15th ACM international conference on Multimedia}.\hskip 1em plus 0.5em minus
  0.4em\relax ACM, 2007, pp. 403--412.

\bibitem{he2004learning}
X.~He, W.-Y. Ma, and H.-J. Zhang, ``Learning an image manifold for retrieval,''
  in \emph{Proceedings of the 12th annual ACM international conference on
  Multimedia}.\hskip 1em plus 0.5em minus 0.4em\relax ACM, 2004, pp. 17--23.

\bibitem{allab2017a}
K.~Allab, L.~Labiod, and M.~Nadif, ``A semi-nmf-pca unified framework for data
  clustering,'' \emph{IEEE Transactions on Knowledge and Data Engineering},
  vol.~29, no.~1, pp. 2--16, 2017.

\bibitem{vandenberghe1996semidefinite}
L.~Vandenberghe and S.~Boyd, ``Semidefinite programming,'' \emph{SIAM Review},
  vol.~38, no.~1, pp. 49--95, 1996.

\bibitem{golub1970singular}
G.~H. Golub and C.~Reinsch, ``Singular value decomposition and least squares
  solutions,'' \emph{Numerische Mathematik}, vol.~14, no.~5, pp. 403--420,
  1970.

\bibitem{ahmed2013distributed}
A.~Ahmed, N.~Shervashidze, S.~Narayanamurthy, V.~Josifovski, and A.~J. Smola,
  ``Distributed large-scale natural graph factorization,'' in \emph{Proceedings
  of the 22nd International Conference on World Wide Web}.\hskip 1em plus 0.5em
  minus 0.4em\relax ACM, 2013, pp. 37--48.

\bibitem{yang2015network}
C.~Yang, Z.~Liu, D.~Zhao, M.~Sun, and E.~Y. Chang, ``Network representation
  learning with rich text information,'' in \emph{International Joint
  Conference on Artificial Intelligence}, 2015, pp. 2111--2117.

\bibitem{xia2019random}
F.~Xia, J.~Liu, H.~Nie, Y.~Fu, L.~Wan, and X.~Kong, ``Random walks: A review of
  algorithms and applications,'' \emph{IEEE Transactions on Emerging Topics in
  Computational Intelligence}, vol.~4, no.~2, pp. 95--107, 2019.

\bibitem{xia2014mvcwalker}
F.~Xia, Z.~Chen, W.~Wang, J.~Li, and L.~T. Yang, ``Mvcwalker: Random walk-based
  most valuable collaborators recommendation exploiting academic factors,''
  \emph{IEEE Transactions on Emerging Topics in Computing}, vol.~2, no.~3, pp.
  364--375, 2014.

\bibitem{al2018analysis}
M.~A. Al-Garadi, K.~D. Varathan, S.~D. Ravana, E.~Ahmed, G.~Mujtaba, M.~U.~S.
  Khan, and S.~U. Khan, ``Analysis of online social network connections for
  identification of influential users: Survey and open research issues,''
  \emph{ACM Computing Surveys (CSUR)}, vol.~51, no.~1, pp. 1--37, 2018.

\bibitem{perozzi2014deepwalk}
B.~Perozzi, R.~Al-Rfou, and S.~Skiena, ``Deepwalk: Online learning of social
  representations,'' in \emph{Proceedings of the 20th ACM SIGKDD International
  Conference on Knowledge Discovery and Data Mining}.\hskip 1em plus 0.5em
  minus 0.4em\relax ACM, 2014, pp. 701--710.

\bibitem{levy2014neural}
O.~Levy and Y.~Goldberg, ``Neural word embedding as implicit matrix
  factorization,'' in \emph{Advances in Neural Information Processing Systems},
  2014, pp. 2177--2185.

\bibitem{rong2014word2vec}
X.~Rong, ``word2vec parameter learning explained,'' \emph{arXiv preprint
  arXiv:1411.2738}, 2014.

\bibitem{goldberg2014word2vec}
Y.~Goldberg and O.~Levy, ``word2vec explained: Deriving mikolov et al.'s
  negative-sampling word-embedding method,'' \emph{arXiv preprint
  arXiv:1402.3722}, 2014.

\bibitem{tang2015www}
J.~Tang, M.~Qu, M.~Wang, M.~Zhang, J.~Yan, and Q.~Mei, ``Line: Large-scale
  information network embedding,'' in \emph{Proceedings of the 24th
  International Conference on World Wide Web}, 2015, pp. 1067--1077.

\bibitem{wang2019sustainable}
W.~Wang, J.~Liu, Z.~Yang, X.~Kong, and F.~Xia, ``Sustainable collaborator
  recommendation based on conference closure,'' \emph{IEEE Transactions on
  Computational Social Systems}, vol.~6, no.~2, pp. 311--322, 2019.

\bibitem{tu2016max}
C.~Tu, W.~Zhang, Z.~Liu, M.~Sun \emph{et~al.}, ``Max-margin deepwalk:
  Discriminative learning of network representation.'' in \emph{International
  Joint Conference on Artificial Intelligence}, 2016, pp. 3889--3895.

\bibitem{ribeiro2017struc2vec}
L.~F. Ribeiro, P.~H. Saverese, and D.~R. Figueiredo, ``struc2vec: Learning node
  representations from structural identity,'' in \emph{Proceedings of the 23rd
  ACM SIGKDD International Conference on Knowledge Discovery and Data
  Mining}.\hskip 1em plus 0.5em minus 0.4em\relax ACM, 2017, pp. 385--394.

\bibitem{yang2016revisiting}
Z.~Yang, W.~W. Cohen, and R.~Salakhutdinov, ``Revisiting semi-supervised
  learning with graph embeddings,'' in \emph{Proceedings of The 33rd
  International Conference on Machine Learning}, 2016, pp. 40--48.

\bibitem{adhikari2017distributed}
B.~Adhikari, Y.~Zhang, N.~Ramakrishnan, and B.~A. Prakash, ``Distributed
  representations of subgraphs,'' in \emph{2017 IEEE International Conference
  on Data Mining Workshops (ICDMW)}.\hskip 1em plus 0.5em minus 0.4em\relax
  IEEE, 2017, pp. 111--117.

\bibitem{narayanan2017graph2vec}
A.~Narayanan, M.~Chandramohan, R.~Venkatesan, L.~Chen, Y.~Liu, and S.~Jaiswal,
  ``graph2vec: Learning distributed representations of graphs,'' \emph{arXiv
  preprint arXiv:1707.05005}, 2017.

\bibitem{benson2017spacey}
A.~R. Benson, D.~F. Gleich, and L.-H. Lim, ``The spacey random walk: A
  stochastic process for higher-order data,'' \emph{SIAM Review}, vol.~59,
  no.~2, pp. 321--345, 2017.

\bibitem{wang2018graphgan}
H.~Wang, J.~Wang, J.~Wang, M.~Zhao, W.~Zhang, F.~Zhang, X.~Xie, and M.~Guo,
  ``Graphgan: Graph representation learning with generative adversarial nets,''
  in \emph{Thirty-Second AAAI Conference on Artificial Intelligence}, 2018, pp.
  2508--2515.

\bibitem{bojchevski2018netgan}
A.~Bojchevski, O.~Shchur, D.~Z{\"u}gner, and S.~G{\"u}nnemann, ``Netgan:
  Generating graphs via random walks,'' \emph{Proceedings of the 35th
  International Conference on Machine Learning (ICML 2018)}, pp. 609--618,
  2018.

\bibitem{shi2017survey}
C.~Shi, Y.~Li, J.~Zhang, Y.~Sun, and S.~Y. Philip, ``A survey of heterogeneous
  information network analysis,'' \emph{IEEE Transactions on Knowledge and Data
  Engineering}, vol.~29, no.~1, pp. 17--37, 2017.

\bibitem{lao2010relational}
N.~Lao and W.~W. Cohen, ``Relational retrieval using a combination of
  path-constrained random walks,'' \emph{Machine learning}, vol.~81, no.~1, pp.
  53--67, 2010.

\bibitem{wang2017knowledge}
Q.~Wang, Z.~Mao, B.~Wang, and L.~Guo, ``Knowledge graph embedding: A survey of
  approaches and applications,'' \emph{IEEE Transactions on Knowledge and Data
  Engineering}, vol.~29, no.~12, pp. 2724--2743, 2017.

\bibitem{lao2011random}
N.~Lao, T.~Mitchell, and W.~W. Cohen, ``Random walk inference and learning in a
  large scale knowledge base,'' in \emph{Proceedings of the Conference on
  Empirical Methods in Natural Language Processing}.\hskip 1em plus 0.5em minus
  0.4em\relax Association for Computational Linguistics, 2011, pp. 529--539.

\bibitem{gardner2013improving}
M.~Gardner, P.~P. Talukdar, B.~Kisiel, and T.~Mitchell, ``Improving learning
  and inference in a large knowledge-base using latent syntactic cues,'' in
  \emph{Proceedings of the 2013 Conference on Empirical Methods in Natural
  Language Processing}, 2013, pp. 833--838.

\bibitem{gardner2014incorporating}
M.~Gardner, P.~Talukdar, J.~Krishnamurthy, and T.~Mitchell, ``Incorporating
  vector space similarity in random walk inference over knowledge bases,'' in
  \emph{Proceedings of the 2014 Conference on Empirical Methods in Natural
  Language Processing (EMNLP)}, 2014, pp. 397--406.

\bibitem{wang2015joint}
W.~Y. Wang and W.~W. Cohen, ``Joint information extraction and reasoning: A
  scalable statistical relational learning approach,'' in \emph{Proceedings of
  the 53rd Annual Meeting of the Association for Computational Linguistics and
  the 7th International Joint Conference on Natural Language Processing (Volume
  1: Long Papers)}, 2015, pp. 355--364.

\bibitem{liu2016hierarchical}
Q.~Liu, L.~Jiang, M.~Han, Y.~Liu, and Z.~Qin, ``Hierarchical random walk
  inference in knowledge graphs,'' in \emph{Proceedings of the 39th
  International ACM SIGIR conference on Research and Development in Information
  Retrieval}.\hskip 1em plus 0.5em minus 0.4em\relax ACM, 2016, pp. 445--454.

\bibitem{fu2017hin2vec}
T.-y. Fu, W.-C. Lee, and Z.~Lei, ``Hin2vec: Explore meta-paths in heterogeneous
  information networks for representation learning,'' in \emph{Proceedings of
  the 2017 ACM on Conference on Information and Knowledge Management}.\hskip
  1em plus 0.5em minus 0.4em\relax ACM, 2017, pp. 1797--1806.

\bibitem{dong2017metapath2vec}
Y.~Dong, N.~V. Chawla, and A.~Swami, ``metapath2vec: Scalable representation
  learning for heterogeneous networks,'' in \emph{Proceedings of the 23rd ACM
  SIGKDD International Conference on Knowledge Discovery and Data Mining},
  2017, pp. 135--144.

\bibitem{hussein2018meta}
R.~Hussein, D.~Yang, and P.~Cudr{\'e}-Mauroux, ``Are meta-paths necessary?:
  Revisiting heterogeneous graph embeddings,'' in \emph{Proceedings of the 27th
  ACM International Conference on Information and Knowledge Management}.\hskip
  1em plus 0.5em minus 0.4em\relax ACM, 2018, pp. 437--446.

\bibitem{wanreinforcement}
G.~Wan, B.~Du, S.~Pan, and G.~Haffari, ``Reinforcement learning based meta-path
  discovery in large-scale heterogeneous information networks,'' in \emph{AAAI
  Conference on Artificial Intelligence}.\hskip 1em plus 0.5em minus
  0.4em\relax AAAI, apr 2020.

\bibitem{shi2019heterogeneous}
C.~Shi, B.~Hu, W.~X. Zhao, and S.~Y. Philip, ``Heterogeneous information
  network embedding for recommendation,'' \emph{IEEE Transactions on Knowledge
  and Data Engineering}, vol.~31, no.~2, pp. 357--370, 2019.

\bibitem{tang2015pte}
J.~Tang, M.~Qu, and Q.~Mei, ``Pte: Predictive text embedding through
  large-scale heterogeneous text networks,'' in \emph{Proceedings of the 21th
  ACM SIGKDD International Conference on Knowledge Discovery and Data
  Mining}.\hskip 1em plus 0.5em minus 0.4em\relax ACM, 2015, pp. 1165--1174.

\bibitem{zhang2019shne}
C.~Zhang, A.~Swami, and N.~V. Chawla, ``Shne: Representation learning for
  semantic-associated heterogeneous networks,'' in \emph{Proceedings of the
  Twelfth ACM International Conference on Web Search and Data Mining}.\hskip
  1em plus 0.5em minus 0.4em\relax ACM, 2019, pp. 690--698.

\bibitem{netembedding2020csr}
M.~Hou, J.~Ren, D.~Zhang, X.~Kong, D.~Zhang, and F.~Xia, ``Network embedding:
  Taxonomies, frameworks and applications,'' \emph{Computer Science Review},
  vol.~38, p. 100296, 2020.

\bibitem{nguyen2018continuous}
G.~H. Nguyen, J.~B. Lee, R.~A. Rossi, N.~K. Ahmed, E.~Koh, and S.~Kim,
  ``Continuous-time dynamic network embeddings,'' in \emph{Companion
  Proceedings of the The Web Conference}, 2018, pp. 969--976.

\bibitem{zuo2018embedding}
Y.~Zuo, G.~Liu, H.~Lin, J.~Guo, X.~Hu, and J.~Wu, ``Embedding temporal network
  via neighborhood formation,'' in \emph{Proceedings of the 24th ACM SIGKDD
  International Conference on Knowledge Discovery \& Data Mining}.\hskip 1em
  plus 0.5em minus 0.4em\relax ACM, 2018, pp. 2857--2866.

\bibitem{hamilton2017inductive}
W.~Hamilton, Z.~Ying, and J.~Leskovec, ``Inductive representation learning on
  large graphs,'' in \emph{Advances in Neural Information Processing Systems},
  2017, pp. 1024--1034.

\bibitem{gori2005new}
M.~Gori, G.~Monfardini, and F.~Scarselli, ``A new model for learning in graph
  domains,'' in \emph{IEEE International Joint Conference on Neural Networks},
  vol.~2.\hskip 1em plus 0.5em minus 0.4em\relax IEEE, 2005, pp. 729--734.

\bibitem{bruna2013spectral}
J.~Bruna, W.~Zaremba, A.~Szlam, and Y.~LeCun, ``Spectral networks and locally
  connected networks on graphs,'' \emph{arXiv preprint arXiv:1312.6203}, 2013.

\bibitem{henaff2015deep}
M.~Henaff, J.~Bruna, and Y.~LeCun, ``Deep convolutional networks on
  graph-structured data,'' \emph{Advances in Neural Information Processing
  Systems}, pp. 1--9, 2015.

\bibitem{defferrard2016convolutional}
M.~Defferrard, X.~Bresson, and P.~Vandergheynst, ``Convolutional neural
  networks on graphs with fast localized spectral filtering,'' in
  \emph{Advances in Neural Information Processing Systems}, 2016, pp.
  3844--3852.

\bibitem{kipf2016semi}
T.~N. Kipf and M.~Welling, ``Semi-supervised classification with graph
  convolutional networks,'' \emph{International Conference on Learning
  Representations}, 2017.

\bibitem{monti2017geometric}
F.~Monti, D.~Boscaini, J.~Masci, E.~Rodola, J.~Svoboda, and M.~M. Bronstein,
  ``Geometric deep learning on graphs and manifolds using mixture model cnns,''
  in \emph{Proceedings of the IEEE Conference on Computer Vision and Pattern
  Recognition}, 2017, pp. 5115--5124.

\bibitem{zhou2017graph}
Z.~Zhou and X.~Li, ``Graph convolution: a high-order and adaptive approach,''
  \emph{arXiv preprint arXiv:1706.09916}, 2017.

\bibitem{manessi2017dynamic}
F.~Manessi, A.~Rozza, and M.~Manzo, ``Dynamic graph convolutional networks,''
  \emph{arXiv preprint arXiv:1704.06199}, 2017.

\bibitem{li2017diffusion}
Y.~Li, R.~Yu, C.~Shahabi, and Y.~Liu, ``Diffusion convolutional recurrent
  neural network: Data-driven traffic forecasting,'' \emph{International
  Conference on Learning Representations}, 2017.

\bibitem{yu2017spatio}
B.~Yu, H.~Yin, and Z.~Zhu, ``Spatio-temporal graph convolutional networks: A
  deep learning framework for traffic forecasting,'' \emph{Proceedings of the
  Twenty-Seventh International Joint Conference on Artificial Intelligence},
  pp. 3634--3640, 2017.

\bibitem{yan2018spatial}
S.~Yan, Y.~Xiong, and D.~Lin, ``Spatial temporal graph convolutional networks
  for skeleton-based action recognition,'' in \emph{Thirty-Second AAAI
  Conference on Artificial Intelligence}, 2018, pp. 3634--3640.

\bibitem{niepert2016learning}
M.~Niepert, M.~Ahmed, and K.~Kutzkov, ``Learning convolutional neural networks
  for graphs,'' in \emph{International Conference on Machine Learning}, 2016,
  pp. 2014--2023.

\bibitem{duvenaud2015convolutional}
D.~K. Duvenaud, D.~Maclaurin, J.~Iparraguirre, R.~Bombarell, T.~Hirzel,
  A.~Aspuru-Guzik, and R.~P. Adams, ``Convolutional networks on graphs for
  learning molecular fingerprints,'' in \emph{Advances in Neural Information
  Processing Systems}, 2015, pp. 2224--2232.

\bibitem{atwood2016diffusion}
J.~Atwood and D.~Towsley, ``Diffusion-convolutional neural networks,'' in
  \emph{Advances in Neural Information Processing Systems}, 2016, pp.
  1993--2001.

\bibitem{gilmer2017neural}
J.~Gilmer, S.~S. Schoenholz, P.~F. Riley, O.~Vinyals, and G.~E. Dahl, ``Neural
  message passing for quantum chemistry,'' in \emph{Proceedings of the 34th
  International Conference on Machine Learning-Volume 70}.\hskip 1em plus 0.5em
  minus 0.4em\relax JMLR. org, 2017, pp. 1263--1272.

\bibitem{allamanis2017learning}
M.~Allamanis, M.~Brockschmidt, and M.~Khademi, ``Learning to represent programs
  with graphs,'' \emph{International Conference on Learning Representations},
  2018.

\bibitem{zhuang2018dual}
C.~Zhuang and Q.~Ma, ``Dual graph convolutional networks for graph-based
  semi-supervised classification,'' in \emph{Proceedings of the Web
  Conference}, 2018, pp. 499--508.

\bibitem{dai2018learning}
H.~Dai, Z.~Kozareva, B.~Dai, A.~Smola, and L.~Song, ``Learning steady-states of
  iterative algorithms over graphs,'' in \emph{International Conference on
  Machine Learning}, 2018, pp. 1114--1122.

\bibitem{aaai20-zhu}
S.~Zhu, L.~Zhou, S.~Pan, C.~Zhou, G.~Yan, and B.~Wang, ``{GSSNN}: Graph
  smoothing splines neural networks,'' in \emph{AAAI Conference on Artificial
  Intelligence}.\hskip 1em plus 0.5em minus 0.4em\relax AAAI, apr 2020.

\bibitem{gao2018large}
H.~Gao, Z.~Wang, and S.~Ji, ``Large-scale learnable graph convolutional
  networks,'' in \emph{Proceedings of the 24th ACM SIGKDD International
  Conference on Knowledge Discovery and Data Mining}.\hskip 1em plus 0.5em
  minus 0.4em\relax ACM, 2018, pp. 1416--1424.

\bibitem{tran2019deepnc}
C.~Tran, W.-Y. Shin, A.~Spitz, and M.~Gertz, ``Deepnc: Deep generative network
  completion,'' \emph{arXiv preprint arXiv:1907.07381}, 2019.

\bibitem{vaswani2017attention}
A.~Vaswani, N.~Shazeer, N.~Parmar, J.~Uszkoreit, L.~Jones, A.~N. Gomez,
  {\L}.~Kaiser, and I.~Polosukhin, ``Attention is all you need,'' in
  \emph{Advances in Neural Information Processing Systems}, 2017, pp.
  5998--6008.

\bibitem{velivckovic2017graph}
P.~Veli{\v{c}}kovi{\'c}, G.~Cucurull, A.~Casanova, A.~Romero, P.~Lio, and
  Y.~Bengio, ``Graph attention networks,'' \emph{International Conference on
  Learning Representations}, 2018.

\bibitem{zhang2018gaan}
J.~Zhang, X.~Shi, J.~Xie, H.~Ma, I.~King, and D.-Y. Yeung, ``{G}a{AN}: Gated
  attention networks for learning on large and spatiotemporal graphs,''
  \emph{Thirty-Fourth Conference on Uncertainty in Artificial Intelligence
  (UAI)}, 2018.

\bibitem{lee2018graph}
J.~B. Lee, R.~Rossi, and X.~Kong, ``Graph classification using structural
  attention,'' in \emph{Proceedings of the 24th ACM SIGKDD International
  Conference on Knowledge Discovery and Data Mining}.\hskip 1em plus 0.5em
  minus 0.4em\relax ACM, 2018, pp. 1666--1674.

\bibitem{abu2018watch}
S.~Abu-El-Haija, B.~Perozzi, R.~Al-Rfou, and A.~A. Alemi, ``Watch your step:
  Learning node embeddings via graph attention,'' in \emph{Advances in Neural
  Information Processing Systems}, 2018, pp. 9180--9190.

\bibitem{hou2021sac}
M.~Hou, L.~Wang, J.~Liu, X.~Kong, and F.~Xia, ``A3graph: Adversarial attributed
  autoencoder for graph representation,'' in \emph{The 36th ACM Symposium on
  Applied Computing (SAC)}, 2021, pp. 1697--1704.

\bibitem{cao2016deep}
S.~Cao, W.~Lu, and Q.~Xu, ``Deep neural networks for learning graph
  representations,'' in \emph{Thirtieth AAAI Conference on Artificial
  Intelligence}, 2016, pp. 1145--1152.

\bibitem{wang2016structural}
D.~Wang, P.~Cui, and W.~Zhu, ``Structural deep network embedding,'' in
  \emph{Proceedings of the 22nd ACM SIGKDD International Conference on
  Knowledge Discovery and Data Mining}.\hskip 1em plus 0.5em minus 0.4em\relax
  ACM, 2016, pp. 1225--1234.

\bibitem{qi2014robust}
Y.~Qi, Y.~Wang, X.~Zheng, and Z.~Wu, ``Robust feature learning by stacked
  autoencoder with maximum correntropy criterion,'' in \emph{2014 IEEE
  International Conference on Acoustics, Speech and Signal Processing
  (ICASSP)}.\hskip 1em plus 0.5em minus 0.4em\relax IEEE, 2014, pp. 6716--6720.

\bibitem{jing2020self}
L.~Jing and Y.~Tian, ``Self-supervised visual feature learning with deep neural
  networks: A survey,'' \emph{IEEE Transactions on Pattern Analysis and Machine
  Intelligence}, 2020.

\bibitem{tu2018deep}
K.~Tu, P.~Cui, X.~Wang, P.~S. Yu, and W.~Zhu, ``Deep recursive network
  embedding with regular equivalence,'' in \emph{Proceedings of the 24th ACM
  SIGKDD International Conference on Knowledge Discovery and Data
  Mining}.\hskip 1em plus 0.5em minus 0.4em\relax ACM, 2018, pp. 2357--2366.

\bibitem{kipf2016variational}
T.~N. Kipf and M.~Welling, ``Variational graph auto-encoders,'' \emph{arXiv
  preprint arXiv:1611.07308}, 2016.

\bibitem{pan2019learning}
S.~Pan, R.~Hu, S.-f. Fung, G.~Long, J.~Jiang, and C.~Zhang, ``Learning graph
  embedding with adversarial training methods,'' \emph{IEEE Transactions on
  Cybernetics}, 2019.

\bibitem{schlichtkrull2018modeling}
M.~Schlichtkrull, T.~N. Kipf, P.~Bloem, R.~Van Den~Berg, I.~Titov, and
  M.~Welling, ``Modeling relational data with graph convolutional networks,''
  in \emph{European Semantic Web Conference}.\hskip 1em plus 0.5em minus
  0.4em\relax Springer, 2018, pp. 593--607.

\bibitem{li2018learning}
Y.~Li, O.~Vinyals, C.~Dyer, R.~Pascanu, and P.~Battaglia, ``Learning deep
  generative models of graphs,'' \emph{arXiv preprint arXiv:1803.03324}, 2018.

\bibitem{xian2018zero}
Y.~Xian, C.~H. Lampert, B.~Schiele, and Z.~Akata, ``Zero-shot learning—a
  comprehensive evaluation of the good, the bad and the ugly,'' \emph{IEEE
  Transactions on Pattern Analysis and Machine Intelligence}, vol.~41, no.~9,
  pp. 2251--2265, 2018.

\bibitem{vyas2020leveraging}
M.~R. Vyas, H.~Venkateswara, and S.~Panchanathan, ``Leveraging seen and unseen
  semantic relationships for generative zero-shot learning,'' in \emph{European
  Conference on Computer Vision}.\hskip 1em plus 0.5em minus 0.4em\relax
  Springer, 2020, pp. 70--86.

\bibitem{wu2019graph}
Z.~Wu, S.~Pan, G.~Long, J.~Jiang, and C.~Zhang, ``Graph wavenet for deep
  spatial-temporal graph modeling,'' in \emph{Proceedings of the 28th
  International Joint Conference on Artificial Intelligence}.\hskip 1em plus
  0.5em minus 0.4em\relax AAAI Press, 2019, pp. 1907--1913.

\bibitem{yao2019graph}
L.~Yao, C.~Mao, and Y.~Luo, ``Graph convolutional networks for text
  classification,'' in \emph{Proceedings of the AAAI Conference on Artificial
  Intelligence}, vol.~33, 2019, pp. 7370--7377.

\bibitem{zhang2018sentence}
Y.~Zhang, Q.~Liu, and L.~Song, ``Sentence-state {LSTM} for text
  representation,'' \emph{The 56th Annual Meeting of the Association for
  Computational Linguistics}, pp. 317--327, 2018.

\bibitem{marcheggiani2017encoding}
D.~Marcheggiani and I.~Titov, ``Encoding sentences with graph convolutional
  networks for semantic role labeling,'' in \emph{Proceedings of the 2017
  Conference on Empirical Methods in Natural Language Processing}, 2017, pp.
  1506--1515.

\bibitem{beck2018graph}
D.~Beck, G.~Haffari, and T.~Cohn, ``Graph-to-sequence learning using gated
  graph neural networks,'' in \emph{Proceedings of the 56th Annual Meeting of
  the Association for Computational Linguistics (Volume 1: Long Papers)}, 2018,
  pp. 273--283.

\bibitem{peng2018large}
H.~Peng, J.~Li, Y.~He, Y.~Liu, M.~Bao, L.~Wang, Y.~Song, and Q.~Yang,
  ``Large-scale hierarchical text classification with recursively regularized
  deep graph-cnn,'' in \emph{Proceedings of the Web Conference}, 2018, pp.
  1063--1072.

\bibitem{wang2018deep}
Z.~Wang, T.~Chen, J.~Ren, W.~Yu, H.~Cheng, and L.~Lin, ``Deep reasoning with
  knowledge graph for social relationship understanding,'' in \emph{Proceedings
  of the 27th International Joint Conference on Artificial Intelligence}.\hskip
  1em plus 0.5em minus 0.4em\relax AAAI Press, 2018, pp. 1021--1028.

\bibitem{lee2018multi}
C.-W. Lee, W.~Fang, C.-K. Yeh, and Y.-C. Frank~Wang, ``Multi-label zero-shot
  learning with structured knowledge graphs,'' in \emph{Proceedings of the IEEE
  Conference on Computer Vision and Pattern Recognition}, 2018, pp. 1576--1585.

\bibitem{battaglia2016interaction}
P.~Battaglia, R.~Pascanu, M.~Lai, D.~J. Rezende \emph{et~al.}, ``Interaction
  networks for learning about objects, relations and physics,'' in
  \emph{Advances in Neural Information Processing systems}, 2016, pp.
  4502--4510.

\bibitem{watters2017visual}
N.~Watters, D.~Zoran, T.~Weber, P.~Battaglia, R.~Pascanu, and A.~Tacchetti,
  ``Visual interaction networks: Learning a physics simulator from video,'' in
  \emph{Advances in Neural Information Processing systems}, 2017, pp.
  4539--4547.

\bibitem{butler2018machine}
K.~T. Butler, D.~W. Davies, H.~Cartwright, O.~Isayev, and A.~Walsh, ``Machine
  learning for molecular and materials science,'' \emph{Nature}, vol. 559, no.
  7715, pp. 547--555, 2018.

\bibitem{rhee2017hybrid}
S.~Rhee, S.~Seo, and S.~Kim, ``Hybrid approach of relation network and
  localized graph convolutional filtering for breast cancer subtype
  classification,'' in \emph{Proceedings of the 27th International Joint
  Conference on Artificial Intelligence}.\hskip 1em plus 0.5em minus
  0.4em\relax AAAI Press, 2018, pp. 3527--3534.

\bibitem{ji2020survey}
S.~Ji, S.~Pan, E.~Cambria, P.~Marttinen, and P.~S. Yu, ``A survey on knowledge
  graphs: Representation, acquisition and applications,'' \emph{arXiv preprint
  arXiv:2002.00388}, 2020.

\bibitem{hamaguchi2018knowledge}
T.~Hamaguchi, H.~Oiwa, M.~Shimbo, and Y.~Matsumoto, ``Knowledge base completion
  with out-of-knowledge-base entities: A graph neural network approach,''
  \emph{Transactions of the Japanese Society for Artificial Intelligence},
  vol.~33, pp. 1--10, 2018.

\bibitem{bordes2013translating}
A.~Bordes, N.~Usunier, A.~Garcia-Duran, J.~Weston, and O.~Yakhnenko,
  ``Translating embeddings for modeling multi-relational data,'' in
  \emph{Advances in Neural Information Processing Systems}, 2013, pp.
  2787--2795.

\bibitem{wang2014knowledge}
Z.~Wang, J.~Zhang, J.~Feng, and Z.~Chen, ``Knowledge graph embedding by
  translating on hyperplanes,'' in \emph{Twenty-Eighth AAAI Conference on
  Artificial Intelligence}, 2014, pp. 1112--1119.

\bibitem{lin2015learning}
Y.~Lin, Z.~Liu, M.~Sun, Y.~Liu, and X.~Zhu, ``Learning entity and relation
  embeddings for knowledge graph completion,'' in \emph{Twenty-ninth AAAI
  conference on artificial intelligence}, 2015, pp. 2181--2187.

\bibitem{ji2015knowledge}
G.~Ji, S.~He, L.~Xu, K.~Liu, and J.~Zhao, ``Knowledge graph embedding via
  dynamic mapping matrix,'' in \emph{Proceedings of the 53rd Annual Meeting of
  the Association for Computational Linguistics and the 7th International Joint
  Conference on Natural Language Processing (Volume 1: Long Papers)}, vol.~1,
  2015, pp. 687--696.

\bibitem{feng2016knowledge}
J.~Feng, M.~Huang, M.~Wang, M.~Zhou, Y.~Hao, and X.~Zhu, ``Knowledge graph
  embedding by flexible translation,'' in \emph{Fifteenth International
  Conference on the Principles of Knowledge Representation and Reasoning},
  2016, pp. 557--560.

\bibitem{huang2017heterogeneous}
Z.~Huang and N.~Mamoulis, ``Heterogeneous information network embedding for
  meta path based proximity,'' \emph{arXiv preprint arXiv:1701.05291}, 2017.

\bibitem{jenatton2012latent}
R.~Jenatton, N.~L. Roux, A.~Bordes, and G.~R. Obozinski, ``A latent factor
  model for highly multi-relational data,'' in \emph{Advances in Neural
  Information Processing Systems}, 2012, pp. 3167--3175.

\bibitem{yang2014embedding}
B.~Yang, W.-t. Yih, X.~He, J.~Gao, and L.~Deng, ``Embedding entities and
  relations for learning and inference in knowledge bases,''
  \emph{International Conference on Learning Representations}, 2015.

\bibitem{liu2017analogical}
H.~Liu, Y.~Wu, and Y.~Yang, ``Analogical inference for multi-relational
  embeddings,'' in \emph{Proceedings of the 34th International Conference on
  Machine Learning-Volume 70}.\hskip 1em plus 0.5em minus 0.4em\relax JMLR.
  org, 2017, pp. 2168--2178.

\bibitem{bello2016neural}
I.~Bello, H.~Pham, Q.~V. Le, M.~Norouzi, and S.~Bengio, ``Neural combinatorial
  optimization with reinforcement learning,'' \emph{International Conference on
  Learning Representations}, 2017.

\bibitem{khalil2017learning}
E.~Khalil, H.~Dai, Y.~Zhang, B.~Dilkina, and L.~Song, ``Learning combinatorial
  optimization algorithms over graphs,'' in \emph{Advances in Neural
  Information Processing Systems}, 2017, pp. 6348--6358.

\bibitem{nowak2018revised}
A.~Nowak, S.~Villar, A.~S. Bandeira, and J.~Bruna, ``Revised note on learning
  quadratic assignment with graph neural networks,'' in \emph{2018 IEEE Data
  Science Workshop (DSW)}.\hskip 1em plus 0.5em minus 0.4em\relax IEEE, 2018,
  pp. 1--5.

\end{thebibliography}

\begin{IEEEbiography}[{\includegraphics[width=1in,height=1.25in,clip,keepaspectratio]{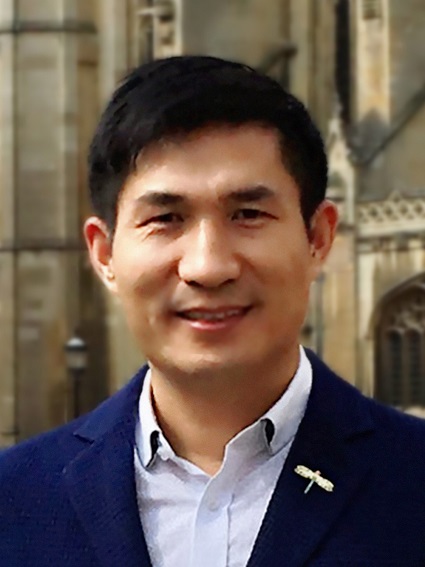}}]{Feng Xia}
(M'07$-$SM'12) received the B.Sc. and Ph.D. degrees from Zhejiang University, Hangzhou, China. He is currently an Associate Professor and Discipline Leader in School of Engineering, IT and Physical Sciences, Federation University Australia. Dr. Xia has published 2 books and over 300 scientific papers in international journals and conferences. His research interests include data science, computational intelligence, social computing, and systems engineering. He is a Senior Member of IEEE and ACM.
\end{IEEEbiography}

\begin{IEEEbiography}[{\includegraphics[width=1in,height=1.25in,clip,keepaspectratio]{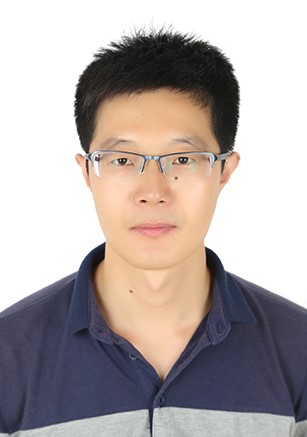}}]{Ke Sun} received the B.Sc. and M.Sc. degrees from Shandong Normal University, Jinan, China. He is currently Ph.D. Candidate in Software Engineering at Dalian University of Technology, Dalian, China. His research interests include deep learning, network representation learning, and knowledge graph.\end{IEEEbiography}

\begin{IEEEbiography}[{\includegraphics[width=1in,height=1.25in,clip,keepaspectratio]{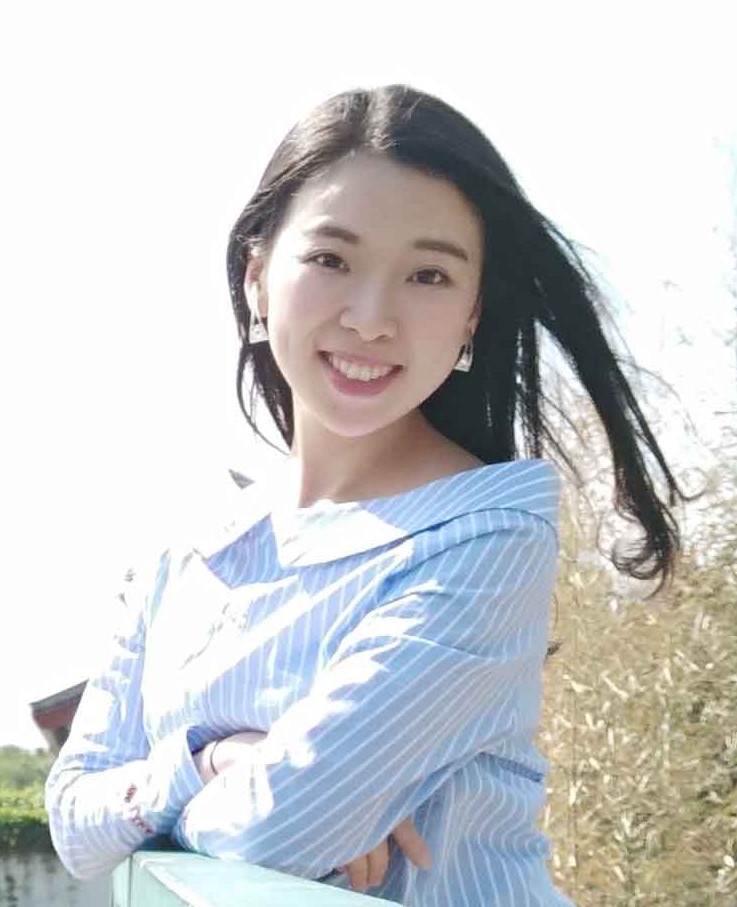}}]{Shuo Yu}(M'20) received the B.Sc. and M.Sc. degrees from Shenyang University of Technology, China, and the Ph.D. degree from Dalian University of Technology, Dalian, China. She is currently a Post-Doctoral Research Fellow with the School of Software, Dalian University of Technology. She has published over 30 papers in ACM/IEEE conferences, journals, and magazines. Her research interests include network science, data
science, and computational social science.\end{IEEEbiography}

\begin{IEEEbiography}[{\includegraphics[width=1in,height=1.25in,clip,keepaspectratio]{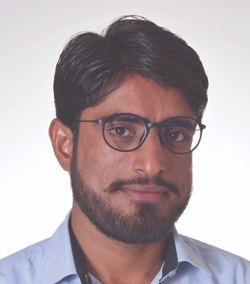}}]{Abdul Aziz} received the Bachelor's degree in computer science from COMSATS Institute of Information Technology, Lahore Pakistan in 2013 and Master degree in Computer science from National University of Computer \& Emerging Sciences Karachi in 2018. He is currently a PhD student at the Alpha Lab, Dalian University of Technology, China. His research interests include big data, information retrieval, graph learning, and social computing.\end{IEEEbiography}

\begin{IEEEbiography}[{\includegraphics[width=1in,height=1.25in,clip,keepaspectratio]{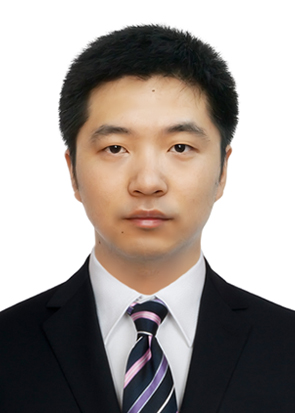}}]{Liangtian Wan}(M'15) received the B.S. degree and the Ph.D. degree from Harbin Engineering University, Harbin, China, in 2011 and 2015, respectively. From Oct. 2015 to Apr. 2017, he has been a Research Fellow at Nanyang Technological University, Singapore. He is currently an Associate Professor of School of Software, Dalian University of Technology, China. He is the author of over 70 papers. His current research interests include data science, big data and graph learning.\end{IEEEbiography}

\begin{IEEEbiography}[{\includegraphics[width=1in,height=1.25in,clip,keepaspectratio]{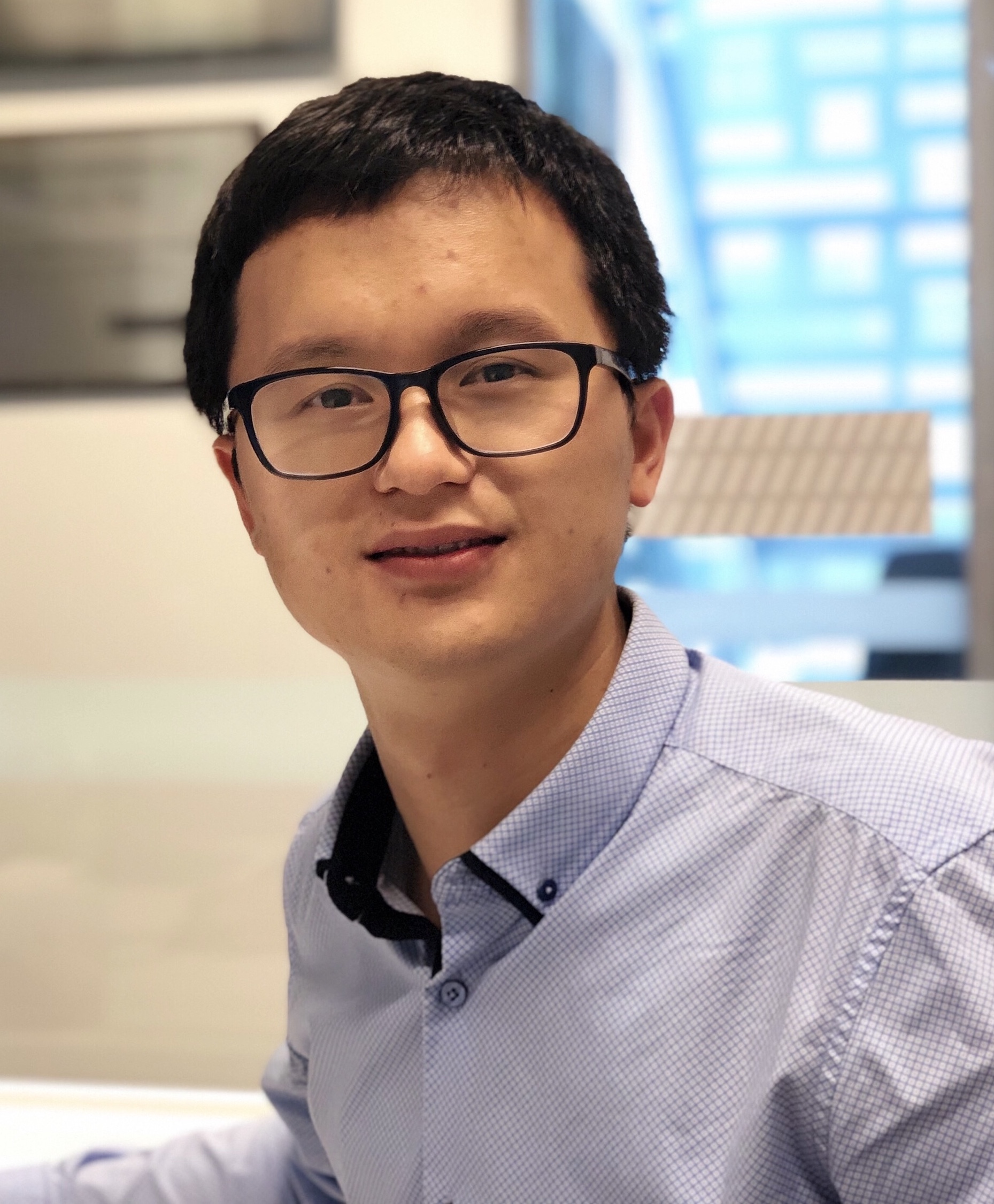}}]{Shirui Pan} received a Ph.D. in computer science from the University of Technology Sydney (UTS), Australia. He is currently a lecturer with the Faculty of Information Technology, Monash University, Australia. His research interests include data mining and machine learning. Dr Pan has published over 60 research papers in top-tier journals and conferences.\end{IEEEbiography}

\begin{IEEEbiography}[{\includegraphics[width=1in,height=1.25in,clip,keepaspectratio]{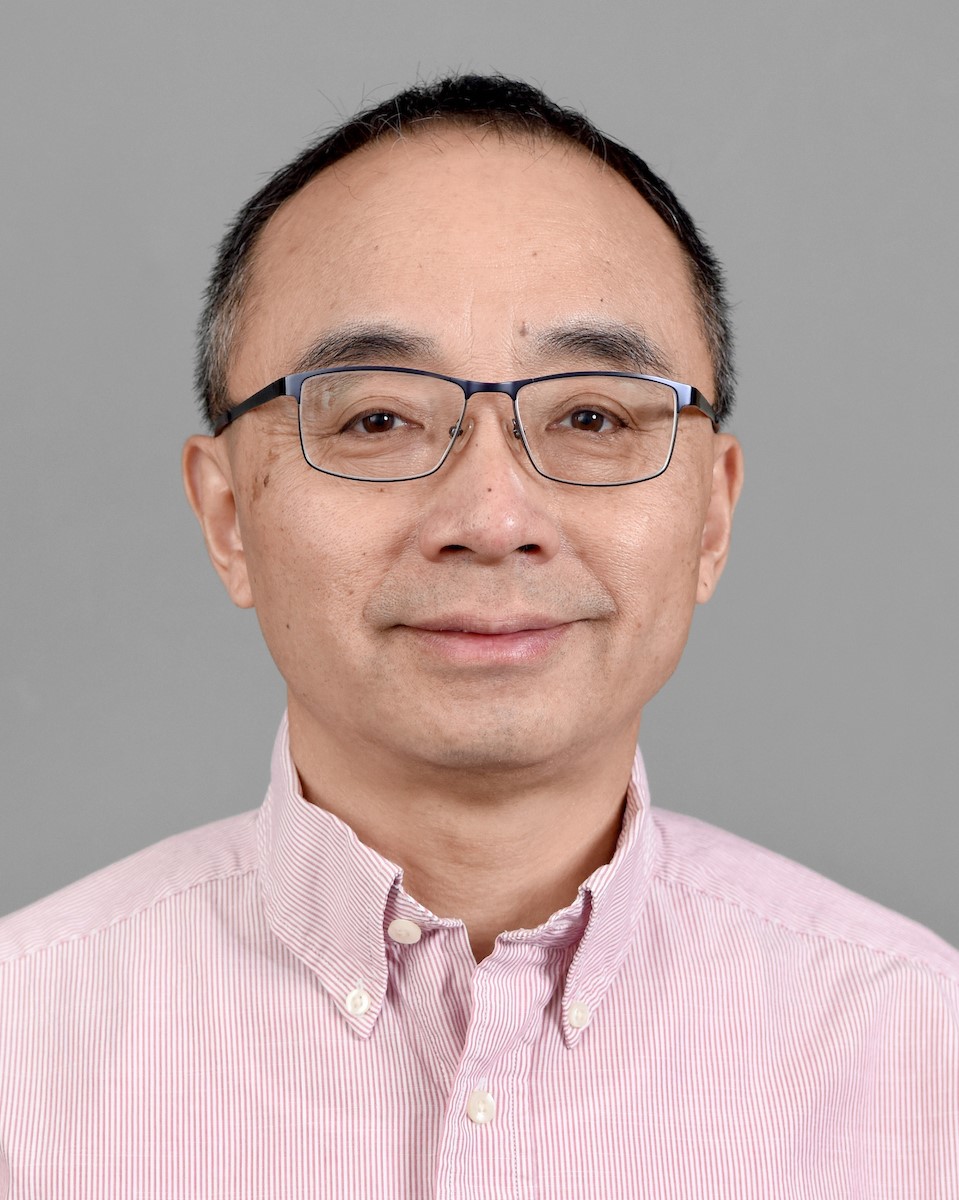}}]{Huan Liu}
(F'12) received the B.Eng. degree in computer science and electrical engineering from Shanghai Jiaotong University and the Ph.D. degree in computer science from the University of Southern California. He is currently a Professor of computer science and engineering at Arizona State University. His research interests include data mining, machine learning, social computing, and artificial intelligence, investigating problems that arise in many real-world applications with high-dimensional data of disparate forms. His well-cited publications include books, book chapters, and encyclopedia entries and conference, and journal papers. He is a Fellow of IEEE, ACM, AAAI, and AAAS.\end{IEEEbiography}

\end{document}